\documentclass[conference]{IEEEtran}
\IEEEoverridecommandlockouts
\usepackage{cite}
\usepackage{amsmath,amssymb,amsfonts}
\usepackage{algorithmic}
\usepackage{graphicx}
\usepackage{textcomp}
\usepackage{xcolor}
\def\BibTeX{{\rm B\kern-.05em{\sc i\kern-.025em b}\kern-.08em
    T\kern-.1667em\lower.7ex\hbox{E}\kern-.125emX}}
\usepackage{booktabs}
\usepackage{listings}
\usepackage{bm}
\usepackage{color}
\usepackage{url}
\usepackage{soul}
\usepackage{CJKulem}
\usepackage{enumitem}
\usepackage{ragged2e}
\usepackage{flushend, cuted}
\usepackage[export]{adjustbox} 
\usepackage{multirow}
\usepackage{amsmath,amssymb}
\usepackage{theorem}
\usepackage{tabularx}
\usepackage{graphicx,subfigure}
\usepackage[ruled,linesnumbered]{algorithm2e}
\usepackage{hyperref}
\usepackage{setspace}
\usepackage{multicol}
\usepackage{graphicx}
\usepackage{stfloats}

\ifCLASSOPTIONcompsoc
  \usepackage[nocompress]{cite}
\else
  \usepackage{cite}
\fi

\begin{document}



 
\title{FinGPT-HPC: Efficient Pretraining and Finetuning Large Language Models for Financial Applications with High-Performance Computing}

\author{Xiao-Yang Liu$^{1}$, Jie Zhang$^{2}$, Guoxuan Wang$^{3}$, Weiqing Tong$^{2}$, and Anwar Walid$^{1}$\\

 $^{1}$Department of Electrical Engineering, Columbia University \\

 \normalsize$^{2}$School of Computer Engineering and Science, Shanghai University \\

  \normalsize$^{3}$Department of Computer Science, Whiting School of Engineering, Johns Hopkins University \\

 \footnotesize{Emails: \texttt{\{XL2427,aie13\}@columbia.edu}, \texttt{\{zhangjie,wqtong\}@shu.edu.cn}, \texttt{gwang69@jhu.com}}
 
 }

\IEEEtitleabstractindextext{%

\begin{abstract}
\justifying
Large language models (LLMs) are computationally intensive. The computation workload and the memory footprint grow quadratically with the dimension (layer width).  Most of LLMs' parameters come from the linear layers of the transformer structure and are highly redundant. These linear layers contribute more than $80\%$ of the computation workload and $99\%$ of the model size. To pretrain and finetune LLMs efficiently, there are three major challenges to address: 1) reducing redundancy of the linear layers; 2) reducing GPU memory footprint; 3) improving GPU utilization when using distributed training.  Prior methods, such as LoRA and QLoRA, utilized low-rank matrices and quantization to reduce the number of trainable parameters and model size, respectively. However, the resulting model still consumes a large amount of GPU memory. In this paper, we present high-performance GPU-based methods that exploit low-rank structures to pretrain and finetune LLMs for financial applications. We replace one conventional linear layer of the transformer structure with two narrower linear layers, which allows us to reduce the number of parameters by several orders of magnitude. By quantizing the parameters into low precision ($8$-bit and $4$-bit), the memory consumption of the resulting model is further reduced.  Compared with existing LLMs, our methods achieve a speedup of $1.3\times$ and a model compression ratio of $2.64\times$ for pretaining without accuracy drop. For finetuning, our methods achieve an average accuracy increase of $6.3 \%$ and $24.0 \%$ in general tasks and financial tasks, respectively, and GPU memory consumption ratio of $6.3\times$. The sizes of our models are smaller than $0.59$ GB, allowing inference on a smartphone.

\end{abstract}


\begin{IEEEkeywords}
LLMs, GPUs, low-rank, high-performance computing, FinGPT
\end{IEEEkeywords}}

\maketitle

\IEEEdisplaynontitleabstractindextext

\IEEEpeerreviewmaketitle

\section{Introduction}\label{sec:introduction}



\begin{figure*}[t]
\subfigure[Method 1]{   
\begin{minipage}{9cm}
\includegraphics[scale=0.14]{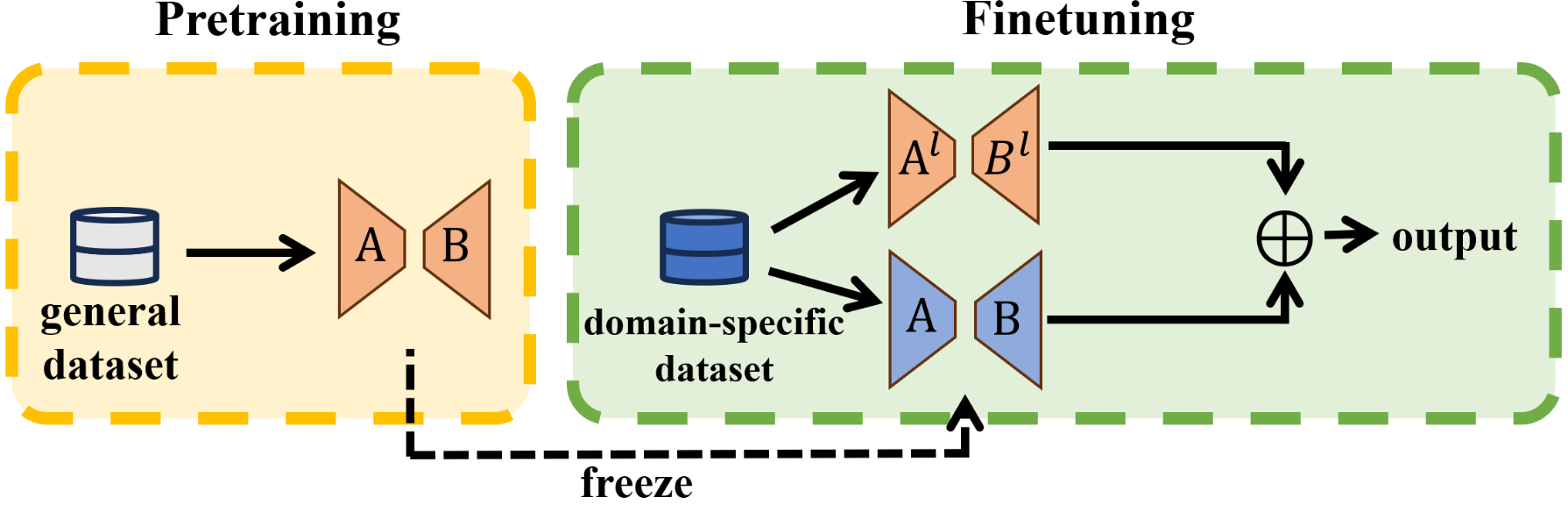}  
\label{fig:lowrank_training}
\end{minipage}
}
\subfigure[Method 2]{ 
\begin{minipage}{9cm}
\includegraphics[scale=0.14]{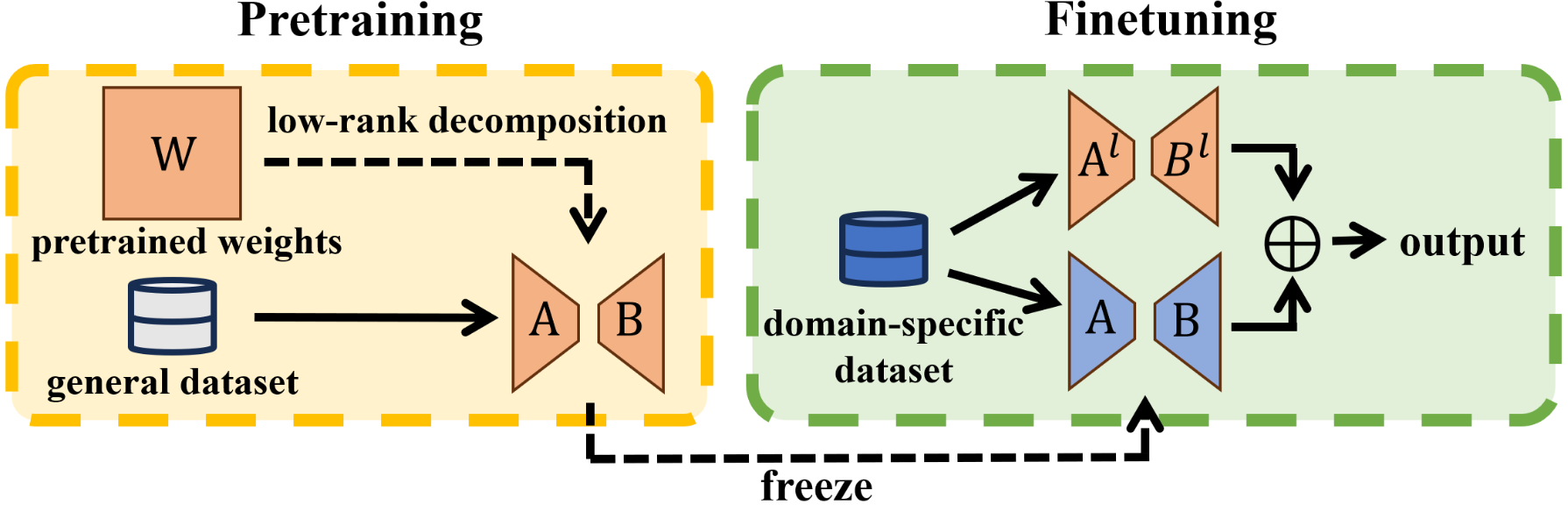}
\label{fig:lowrank_training2}
\end{minipage}
}
\caption{Our methods for pretraining (left) and finetuning (right) LLMs. We replace a linear layer of the transformer network \cite{vaswani2017attention} with two linear layers with weight matrices $\bm{A}$ and $\bm{B}$.}
\label{fig:lowrankpretraining}
\end{figure*}

Large language models (LLMs) have achieved great success in many domains. Pretraining LLMs with hundreds of billions of parameters may take millions of GPU hours, say run $500$ GPUs for $6$ months. Table \ref{table:llm_pandt} provides a summary of the model size and typical training time for popular LLMs. 
A widely adopted approach is finetuning a pretrained LLM on private datasets. However, LLMs have a large number of parameters, and the finetuning process may still consume hundreds of gigabytes of GPU memory. Most of the parameters of LLMs, i.e. $\ge 99\%$, come from the linear layers of the transformer structure \cite{vaswani2017attention}, where the computation workload and the memory footprint grow quadratically with the dimensions (i.e., layer width). For example, the dimensions of hidden layers of Llama2-7B \cite{llama2} and Llama2-13B \cite{llama2} are $4096$ and $5120$, respectively, and the numbers of (decoder) layers are $32$ and $40$, respectively. The model size and training time of Llama2-13B is about twice that of Llama2-7B, as shown in Table \ref{table:llm_pandt}.


\textbf{Challenges}. Employing LLMs is highly computing-intensive and there are several challenges to contend with:
\begin{itemize}
    \item The pretraining dataset is too large to fit into GPU memory (e.g., $32$ GB or $80$ GB) or even CPU memory (e.g., $63$ GB or $128$ GB), as shown in Table \ref{table:llm_pandt}. During the training process, data is loaded into GPU memory from the disk (e.g., $1$ TB or $4$ TB), which is time-intensive.
    \item  LLMs require large GPU memory during both the training and inference stages (more details can be found in Table \ref{table:GPUmemory}).
    \item When using the model-parallel method\cite{krizhevsky2014one} to pretrain an LLM across multiple GPUs, these GPUs remain under low utilization.
\end{itemize}




 The low-rank adaptation (LoRA) method \cite{hulora} is a parameter-efficient finetuning method. It utilizes the low-rank structure to reduce the redundancy of the linear layers in the transformer structure \cite{vaswani2017attention}. Further, the quantized LoRA (QLoRA) method \cite{dettmers2023qlora} converts the precicion of parameters from half-precision ($16$-bit) into  INT8 ($8$-bit) or INT4 ($4$-bit), which can reduce the GPU memory consumption. The TensorGPT method \cite{xu2023tensorgpt} replaces the embedding layer with low-rank tensor-train structure to efficiently capture the weights correlations.


\begin{table}
\caption{Popular LLMs: number of parameters (billion), model size (memory), dataset (disk),  and training time (GPU$\times$hours).}
\centering
\resizebox{\linewidth}{!}{
\begin{tabular}{ccccc}
\toprule
LLMs & \#Parameters & Size (GB) & Data (TB) & Training (hours) \\
\midrule
GPT-3 \cite{GPT3} & $175$B & $350$ & $0.9$ & $835584$  \\
Llama2\cite{llama2} & $7$B; $13$B; $70$B & $14; 26; 140$ & $4.6$ &  $184320; 368640; 1720320$  \\
Falcon \cite{falcon} & $7$B; $40$B; $180$B & $14; 80; 360$ & $2.1$ & $129024; 571392; \sim7000000$ \\
ChatGLM3 \cite{du2022glm} & $6$B & $12$ & $1.3$ & $1105920$  \\
Mistral-$8\times7$B \cite{jiang2023mistral} & $46.7$B & $93.4$ & - & -  \\

\bottomrule
\end{tabular}
}
\label{table:llm_pandt}
\end{table}

In this paper, we present efficient pretraining and finetuning LLMs for financial applications with high-performance computing. Our method has two parts: pretraining and finetuning, as shown in Fig. \ref{fig:lowrankpretraining}. First, we pretrain a low-rank LLM on a general dataset and obtain a general LLM. Second, we finetune the LLM on domain-specific datasets by freezing the pretrained weights and adding a low-rank version on a parallel path.

Our contributions can be summarized as follows.

\begin{itemize}




\item We provide a new training paradigm. We employ the low-rank structure and quantization technique in the linear layers of the transformer structure, leading to a substantial reduction in the number of trainable parameters. Therefore, both the GPU memory footprint and running time of both pretraining and finetuning LLMs are greatly reduced. Furthermore, the response time and memory required in the inference stage are also reduced.

\item We utilize a recomputing technique to reduce the peak number of intermediate variables so that the GPU memory footprint of pretraining and finetuning LLMs are greatly reduced. Using the pipeline model-parallel method, we improve the GPU utilization in pretraining across multiple GPUs, thus reducing the pretraining time.

\item We perform extensive experiments to evaluate the performance of our methods. For pretraining GPT2-1.5B\cite{GPT2}, our method achieves a speedup of $1.3 \times$, and our pretrained LLM has a compression ratio of $2.64 \times$ without accuracy drop. For finetuning Llama2-13B\cite{llama2}, our method achieves an average accuracy increase of $6.3 \%$ and $24.0 \%$ in general tasks and financial tasks, respectively, and a reduction of GPU memory footprint by $6.3 \times$.

\end{itemize}

The remainder of this paper is organized as follows. Section \ref{section:Relate_Works} discusses related works. Section \ref{section:OverviewofTensorLayer} gives a background on LLMs and the low-rank structure. Section \ref{sec:pre-training} presents our method for pretraining LLMs. Section \ref{sec:fine-tuning} presents our method for finetuning LLMs. Section \ref{section:performace} describes the experimental settings and results. In Section \ref{section:ConclusionandFutureWork}, we provide concluding remarks.




\section{Related Works}
\label{section:Relate_Works}

\subsection{Large Language Models (LLMs)}
The transformer structure\cite{vaswani2017attention} uses the self-attention mechanism and allows the model to process sequential data efficiently. It performs much better in processing texts than previous neural network structures. GPT-1\cite{GPT1} is one of the first generative pretrained transformer (GPT). GPT-3\cite{GPT3} has significantly increased the model size from $1.5$ B to $175$ B. It is the base model of ChatGPT, which was trained using instruction finetuning\cite{instructGPT}. LLaMA\cite{touvron2023llama} is an open-source pretrained LLM, which has three versions with parameter sizes of 7B, 13B and 70B, respectively.

J. Kaplan  \textit{et al.} \cite{kaplan2020scaling} proposed the scaling law of LLMs, where the model performance scales as a power-law with model size, dataset size, and the amount of compute used for pretraining. K. Cobbe \textit{et al.} \cite{cobbe2021training} showed that the scaling law also applies to model fine-tuning. J. Wei \textit{et al.} \cite{wei2022emergent}  discovered that LLMs have emergent abilities, which are not present in smaller models but are present in larger models.

\subsection{Low-rank Structure and Quantization Technique}


\textbf{Low-rank structure}. Using the low-rank structure can greatly reduce the parameters of the linear layers. The Low-rank adaptation (LoRA) method\cite{hulora} reduced the number of trainable parameters by a factor of $10,000$ times. However, it still retained a full-size pretrained LLM for inference. Much research focused on decomposing pretrained models \cite{novikov2015tensorizing}, and there was a significant performance drop for low-rank models \cite{hayashi2019exploring}. Tensor layers are suited for computing on GPUs\cite{zhang2019cutensor,lu2019high,li2019high,zhang2021high}, and have the low-rank feature\cite{zhang2019cutensor-tubal,zhang2020high}. H. Huang \textit{et al.} \cite{HT_tensor} used hierarchical tucker layers to reduce the model's accuracy drop, but incurred a longer training time. X.-Y. Liu \textit{et al.} \cite{CP_tensor} used high-performance tucker layers to accelerate the training. 

\textbf{Quantization technique}. Quantization maps high-precision data to low-precision data \cite{dettmers2022gpt3}, which can effectively reduce the memory consumption.
T. Dettmers \textit{et al.} \cite{llmint8} implemented a selective quantization strategy. Data in the outlier vectors is kept in the original precision, while data in other vectors is quantized into \textsf{int8} format. G. Xiao \textit{et al.} \cite{xiao2023smoothquant} discovered that outliers make the activations difficult to quantize and have less impact on the weights. The authors altered the value range of activations and weights, balancing their quantitative difficulty. Besides reducing the effect of outliers, E. Frantar \textit{et al.} \cite{frantar2022gptq} reduced accuracy loss by improving the rounding strategy. However, they did not use quantization in the pretraining stage.



\section{Background on LLMs and Low-rank Structures}
\label{section:OverviewofTensorLayer}

\subsection{Decoder-only Transformer Structure}
\label{sec:transformer}

\begin{figure*}
\centering
\includegraphics[width=18cm]{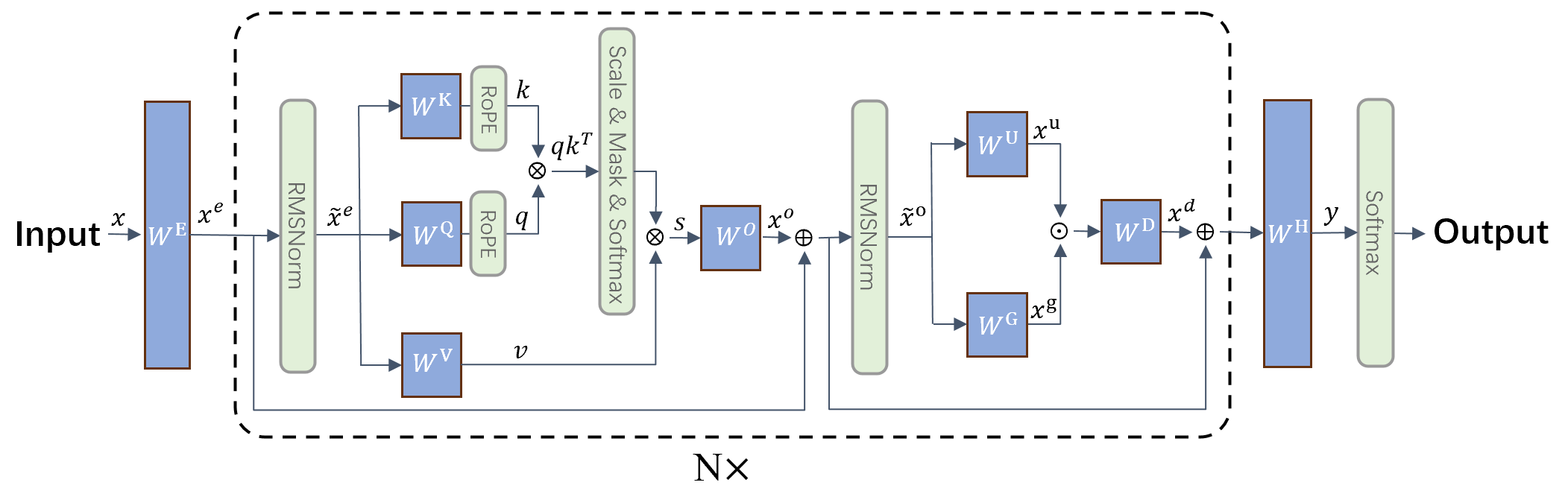}
\caption{The transformer structure (decoder-only) of Llama2.}
\label{fig:transformer_architecture}
\end{figure*}

Most existing LLMs are based on the decoder-only transformer structure \cite{vaswani2017attention}. Fig.  \ref{fig:transformer_architecture} shows the structure of Llama2 \cite{llama2}. It consists of three parts:
\begin{itemize}
    \item First, there is an embedding layer that converts the input $\bm{x}$ to a latent vector $\bm{x}^e$.
    \item Second, there are $N$ decoder layers. Each decoder has two modules: the multi-head self-attention module and the feed-forward network module. The self-attention module transforms $\bm{x}^e$ to $\bm{x}^o$. The feed-forward network module transforms $\bm{x}^o$ to $\bm{x}^d$. In addition, the inputs $\bm{x}^e$ and $\bm{x}^o$ are skip-connected to the output $\bm{x}^o$ and $\bm{x}^d$, respectively. For each module, there is a normalization layer that normalizes $\bm{x}^e$ and $\bm{x}^o$ to $\Tilde{\bm{x}}^{e}$ and $\Tilde{\bm{x}}^{o}$, respectively.
    \item Finally, there is an output layer $\bm{W}^{H}$ that converts the latent representations $\bm{x}^{d}$ to $\bm{y}$. Then, the softmax function transforms $\bm{y}$ to a vector of probabilities, where each entry is the probability of picking the corresponding token.
\end{itemize}

\begin{table*}
  \caption{Breakdown of parameters in \textcolor{blue}{Llama2-7B} ($N=32$) and \textcolor{orange}{Llama2-13B} ($N=40$).}
  \centering
  \begin{tabular}{ccccc}
\toprule
\textbf{Module}&\textbf{Size}& \textbf{Amount (M)} & \textbf{Storage (GB)} & \textbf{Percentage (\%)}\\
\midrule
$\bm{W}^{E}$ & \textcolor{blue}{$32000 \times 4096$}, \textcolor{orange}{$32000 \times 5120$} & \textcolor{blue}{$131.07$}, \textcolor{orange}{$163.84$} & \textcolor{blue}{$0.26$}, \textcolor{orange}{$0.33$} & \textcolor{blue}{1.94}, \textcolor{orange}{1.25} \\
\midrule
RMSNorm & \textcolor{blue}{$4096$}, \textcolor{orange}{$5120$} & \textcolor{blue}{$0.13$}, \textcolor{orange}{$0.21$} & $\ll 0.01$ & $\ll 0.01$ \\
$\bm{W}^{Q}$ & \textcolor{blue}{$4096 \times 4096$}, \textcolor{orange}{$5120 \times 5120$}& \textcolor{blue}{$536.87$}, \textcolor{orange}{$1048.56$} & \textcolor{blue}{$1.07$}, \textcolor{orange}{$2.10$} & \textcolor{blue}{7.98}, \textcolor{orange}{8.07} \\

$\bm{W}^{K}$ & \textcolor{blue}{$4096 \times 4096$}, \textcolor{orange}{$5120 \times 5120$} & \textcolor{blue}{$536.87$}, \textcolor{orange}{$1048.56$} & \textcolor{blue}{$1.07$}, \textcolor{orange}{$2.10$} & \textcolor{blue}{7.98}, \textcolor{orange}{8.07} \\

$\bm{W}^{V}$ & \textcolor{blue}{$4096 \times 4096$}, \textcolor{orange}{$5120 \times 5120$} & \textcolor{blue}{$536.87$}, \textcolor{orange}{$1048.56$} & \textcolor{blue}{$1.07$}, \textcolor{orange}{$2.10$} & \textcolor{blue}{7.98}, \textcolor{orange}{8.07} \\

$\bm{W}^{O}$& \textcolor{blue}{$4096 \times 4096$}, \textcolor{orange}{$5120 \times 5120$} & \textcolor{blue}{$536.87$}, \textcolor{orange}{$1048.56$} & \textcolor{blue}{$1.07$}, \textcolor{orange}{$2.10$} & \textcolor{blue}{7.98}, \textcolor{orange}{8.07} \\

RMSNorm & \textcolor{blue}{$4096$}, \textcolor{orange}{$5120$} & \textcolor{blue}{$0.13$}, \textcolor{orange}{$0.21$} & $\ll 0.01$ & $\ll 0.01$ \\
$\bm{W}^{U}$ & \textcolor{blue}{$4096 \times 11008$}, \textcolor{orange}{$5120 \times 13824$} & \textcolor{blue}{$1442.84$}, \textcolor{orange}{$2834.84$}& \textcolor{blue}{$2.89$}, \textcolor{orange}{$5.67$} & \textcolor{blue}{21.40}, \textcolor{orange}{21.74} \\

$\bm{W}^{G}$ & \textcolor{blue}{$4096 \times 11008$}, \textcolor{orange}{$5120 \times 13824$} &  \textcolor{blue}{$1442.84$}, \textcolor{orange}{$2834.84$}& \textcolor{blue}{$2.89$}, \textcolor{orange}{$5.67$} & \textcolor{blue}{21.40}, \textcolor{orange}{21.74} \\

$\bm{W}^{D}$ & \textcolor{blue}{$11008 \times 4096$}, \textcolor{orange}{$5120 \times 13824$} & \textcolor{blue}{$1442.84$}, \textcolor{orange}{$2834.84$} & \textcolor{blue}{$2.89$}, \textcolor{orange}{$5.67$} & \textcolor{blue}{21.40}, \textcolor{orange}{21.74} \\
\midrule
$\bm{W}^{H}$ & \textcolor{blue}{$4096 \times 32000$}, \textcolor{orange}{$32000 \times 5120$} & \textcolor{blue}{$1442.84$}, \textcolor{orange}{$163.84$} & \textcolor{blue}{$0.26$}, \textcolor{orange}{$0.33$} & \textcolor{blue}{1.94}, \textcolor{orange}{1.25} \\

\bottomrule
\end{tabular}
\label{table:parameters}
\end{table*}
In the multi-head self-attention module, $\Tilde{\bm{x}}^{e}$ is linearly projected $h$ times with different, learned projections to $\bm{k}, \bm{q}, \bm{v} \in \mathbb{R}^{d}$. The number of heads is $h$. Attention functions are performed in parallel. In the attention function, $\bm{qk}^T$ is masked and scaled by $\frac{1}{\sqrt{d}}$ before feeding into the the softmax function. After the attention function, we get $head_i,~i = 1, 2, ...,h$.
\begin{equation}
\label{eq:attention}
\begin{aligned}
\operatorname{\bm{head_i}} =\operatorname{Attention}(\bm{q, k, v})=\operatorname{softmax}(\frac{\bm{q} \bm{k}^{T}}{\sqrt{d}}) \bm{v}, \\
\bm{q}  = \text{RoPE}(\bm{W}_{i}^{Q}\Tilde{\bm{x}}^{e}), ~\bm{k} = \text{RoPE}(\bm{W}_{i}^{K}\Tilde{\bm{x}}^{e}), ~\bm{v} = \bm{W}_{i}^{V}\Tilde{\bm{x}}^{e},
\end{aligned}
\end{equation}
where RoPE \cite{RoPE} is the Rotary Position Embedding function.

Concatenating these $head_i$, input to $\bm{W}^o$ and obtain $\bm{x}^o$:
\begin{equation}
\label{eq:concat}
\bm{x^o} =\operatorname{Concat}\left(\operatorname{head}_{1}, \ldots, \operatorname{head}_{\mathrm{h}}\right) \bm{W}^{O}.
\end{equation}
The feed-forward network consists of three linear layers:
\begin{equation}
\label{eq:FFN}
    \bm{x^{d}} =  \bm{W}^{D}(\underbrace{\bm{W}^{U}\Tilde{\bm{x}}^{o}}_{\bm{x}^u} \odot \operatorname{SiLU}(\underbrace{\bm{W}^{G}\Tilde{\bm{x}}^{o}}_{\bm{x}^{g}})),
\end{equation}
where SiLU is the activation function and $\odot$ is the element-wise product.

Given an input token, $\bm{x} \in \mathbb{R}^{32000}$, the forward processing of Fig. \ref{fig:transformer_architecture} with parameters in Table \ref{table:parameters} is as follows:
\begin{itemize}
    \item Input $\bm{x}$ to an embedding layer $\bm{W}^{E} \in \mathbb{R}^{32000 \times 4096}$, we get an embedding vector $\bm{x}^{e} = \bm{W}^{E}\bm{x} \in \mathbb{R}^{4096}$.
    \item Renormalize $\bm{x}^e \in \mathbb{R}^{4096}$ through an RMSNorm layer and get $\Tilde{\bm{x}}^{e} \in \mathbb{R}^{4096}$.
    \item Perform products of $\Tilde{\bm{x}}^{e}$ with weight matrices $\bm{W}^{K} \in \mathbb{R}^{4096 \times 4096}, \bm{W}^{Q} \in \mathbb{R}^{4096 \times 4096}, \bm{W}^{V} \in \mathbb{R}^{4096 \times 4096}$, the results $\bm{k}, \bm{q}, \bm{v} \in \mathbb{R}^{4096}$ are split into $h = 32$ parts. Each part performs an attention function in parallel as in (\ref{eq:attention}). We get $head_i \in \mathbb{R}^{4096 \times 128}, ~i=1,2,..., 32$.
    \item Concatenate these $head_i$ and performing products with weight matrix $\bm{W}^{O} \in \mathbb{R}^{4096 \times 4096}$, and get $\bm{x}^{o} \in \mathbb{R}^{4096}$ as in (\ref{eq:concat}).
    \item Input $\bm{x}^{o}$ to an RMSNorm layer, we get $\Tilde{\bm{x}}^{o} \in \mathbb{R}^{4096}$.
    \item Perform products of $\Tilde{\bm{x}}^{o}$ with weight matrices $\bm{W}^{G} \in \mathbb{R}^{4096 \times 11008}$ and $\bm{W}^{U} \in \mathbb{R}^{4096 \times 11008}$ in parallel. The result $\bm{x}^g \in \mathbb{R}^{11008}$ performs the dot-product with result $\bm{x}^u \in \mathbb{R}^{11008}$ after SiLU function. The output performs the product with weight $\bm{W}^{D} \in \mathbb{R}^{11008 \times 4096}$ as in (\ref{eq:FFN}). We then get $\bm{x}^{d} \in \mathbb{R}^{4096}$.
    \item Finally, the result of $\bm{x}^{d} + \bm{x}^{o}$ is fed into the output layer $\bm{W}^{H} \in \mathbb{R}^{4096 \times 32000}$, we get $\bm{y} \in \mathbb{R}^{32000}$ and then feed it into the softmax function.
\end{itemize}


\subsection{Low-rank Structure}
\label{sec:lowrank_structure}

As described in Section \ref{sec:transformer}, the linear layers are major building blocks of the transformer network \cite{vaswani2017attention}. An input sample $\bm{x} \in \mathbb{R}^{n}$ is multiplied by a weight matrix $\bm{W} \in \mathbb{R}^{n \times n}$, and the output $\bm{y} \in \mathbb{R}^{n}$ can be represented as follows:
\begin{equation}
\label{eq:linear_layer}
\bm{y} =\bm{W}\bm{x}.
\end{equation} 


Empirical study found that the linear layers are highly redundant \cite{cheng2015exploration},
allowing us to replace $\bm{W} \in \mathbb{R}^{n \times n}$ with two sublinear layers, $\bm{A} \in \mathbb{R}^{r \times n}$ and $\bm{B} \in \mathbb{R}^{n \times r}$ with $ r \ll n$. Then, (\ref{eq:linear_layer}) becomes:
\begin{equation}
\label{eq:low_rank_layer}
\bm{y} =\bm{B}\bm{A} \bm{x}.
\end{equation}
Note that the layer size is reduced from $n^2$ to $2n r$, while the number of multiplications is reduced from $n^2$ to $2n r$, too.

As an example, we replace the weight matrices of Llama2-7B \cite{llama2} with a low-rank structure with $r=512$. $\bm{W}^{K}$, $\bm{W}^{Q}$, $\bm{W}^{V}$, and $\bm{W}^{O}$ are replaced with a pair of low-rank matrices $\bm{A} \in \mathbb{R}^{4096 \times 512}$ and $\bm{B} \in \mathbb{R}^{512 \times 4096}$, respectively. The parameters of each weight matrix are reduced from $16.78$ M to $4.19$ M, and the total parameters of these weight matrices in the model are reduced from $2.15$ B to $0.54$ B. $\bm{W}^{U}$ and $\bm{W}^{G}$ are replaced with a pair of low-rank matrices $\bm{A} \in \mathbb{R}^{4096 \times 512}$ and $\bm{B} \in \mathbb{R}^{512 \times 11008}$, respectively, and $\bm{W}^{D}$ is replaced with a pair of low-rank matrices $\bm{A} \in \mathbb{R}^{11008 \times 512}$ and $\bm{B} \in \mathbb{R}^{512 \times 4096}$. The parameters of each weight matrix are reduced from $45.09$ M to $7.73$ M, and the total parameters of these weight matrices in the model are reduced from $4.33$ B to $0.74$ B. Similarly, the parameters of $\bm{W}^E$ and $\bm{W}^H$ can be reduced from $262.14$ M to $18.48$ M, using such a low-rank structure.

\subsection{Challenges in Training and Inference}
\label{sec:challenges}

\subsubsection{Challenges in Training}
\label{sec:chaInTrain}
\begin{table}
\caption{Breakdown of GPU memory footprint for training Llama2-7B\cite{llama2} and Llama2-70B\cite{llama2} with different batch sizes.}
\Centering
\resizebox{\linewidth}{!}{
\begin{tabular}{cccccc}
\toprule
LLMs & Batch size & Para. (GB) & Grad. (GB)& Opt. (GB)& Inter. (GB)\\
\midrule
\multirow{3}{*}{7B} & 1 & $14$ & $14$ & $84$ & $81$  \\
& 4 & $14$ & $14$ & $84$ & $388$\\
& 16 & $14$ & $14$ & $84$ & $1552$\\
\midrule
\multirow{3}{*}{70B} & 1 & $140$ & $140$ & $840$ &  $442.5$ \\
& 4  & $140$ & $140$ & $840$ &   $1770$ \\
& 16 & $140$ & $140$ & $840$ &   $7080$ \\
\bottomrule
\end{tabular}
}
\label{table:GPUmemory}
\end{table}

LLMs have a large number of parameters and usually take a long training time, as given in Table \ref{table:llm_pandt}. Pretraining an LLM has three primary challenges: 
\begin{itemize}
    \item Large scale dataset is needed to pretrain LLMs. For example, the dataset used to pretrain Llama2 \cite{llama2} is $4.7$ TB, which requires large amounts of GPU and CPU memory.
    \item The memory footprint of pretraining and finetuning exceeds the GPU memory capacity. The GPU memory footprint can be classified into four parts, as shown in Table \ref{table:GPUmemory}. For pretraining Llama2-7B and Llama2-70B with batch size $16$, the GPU memory footprint has a size of $1,664$ GB and $8200$ GB, respectively.
    \item In the model-parallel method, an LLM is split over multiple GPUs. During the pretraining, the forward pass is computed layer by layer and only one GPU is working at a time. When splitting an LLM into $N$ GPUs, each GPU has a utilization of $\frac{1}{N}$, which is relatively low.
\end{itemize}





\subsubsection{Challenges in Inference}
\label{sec:chaInInfer} 
During inference, we load the trained LLM into GPU. Using a standard tokenizer, the input text is processed into a sequence of tokens, $\bm{X} \in \mathbb{R}^{t \times l}$, where $l$ is the sequence length and $t$ is the vocabulary size. The input $\bm{X}$ is then embedded into $\bm{X}^e \in \mathbb{R}^{n \times l}$.
For $h$ heads in multi-head self-attention module, there is an attention function in (\ref{eq:attention}). This function uses $\bm{K}_i, \bm{Q}_i, \bm{V}_i \in \mathbb{R}^{\frac{n}{h} \times l}$ to calculate attention score $\text{softmax}(\frac{\bm{Q}_i\bm{K}_i^T}{\sqrt{d}}) \in \mathbb{R}^{l \times l}$ and $head_i = \text{softmax}(\frac{\bm{Q}_i\bm{K}_i^T}{\sqrt{d}})\bm{V}_i \in \mathbb{R}^{\frac{n}{h} \times l}$, $i = 1, 2, ..., h$. Concatenating these heads, we get the $s \in \mathbb{R}^{n \times l}$ in Fig. \ref{fig:transformer_architecture} to continue the process as Section \ref{sec:transformer}. Finally, the last column $\bm{y} \in \mathbb{R}^{t}$ of output $\bm{Y} \in \mathbb{R}^{ t \times l}$ is the predicted token. 

The inference stage has challenges as follows: 
\label{sec:challangeininfer}
\begin{itemize}
    \item Limited memory capacity poses a challenge in using LLMs for various applications, such as inference on mobile phones. For example, Llama2-7B has a size of $14$ GB as shown in Table \ref{table:GPUmemory}. Using the low-rank structure with $r = 512$, the model has about $2$ billion parameters, which has a size of $4$ GB. It is still too large for mobile phones.
    \item As shown in Fig. \ref{fig:transformer_architecture}, most parameters of the transformer structure come from the linear layers (\ref{eq:linear_layer}). In the inference stage, each parameter requires two operations, multiplication and addition. Therefore, Llama2-7B has a computational workload of $14$ GFLOPS for each token. For generative LLM, each generated token needs an inference, that token is added to the end of input tokens as a new input for the next inference. For example, if the input has $100$ tokens and the LLM generates $100$ tokens, there are $100$ times inference processes, these processes have $100, 101,...,199$ tokens as input, respectively. Therefore, the whole inference stage has $14950$ tokens to be computed. The computational workload of Llama2-7B is $210$ TFLOPS. However, the mobile phone has a computational power of about 2 TFLOPs, and it will take $105$ seconds which does not satisfy the requirement.
\end{itemize}

\begin{figure*}[htbp]
\subfigure[model-parallel]{   
\begin{minipage}{9cm}
\includegraphics[scale=0.17]{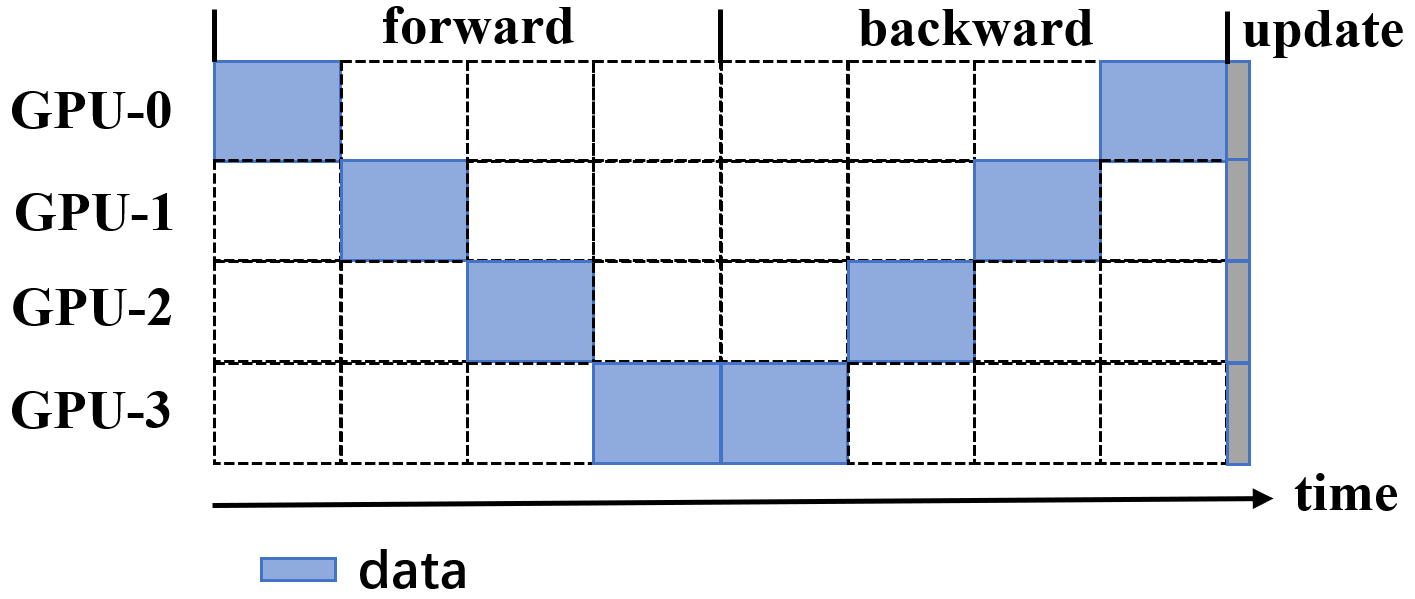}  
\label{fig:modelparallel}
\end{minipage}
}
\subfigure[pipeline model-parallel]{ 
\begin{minipage}{9cm}
\includegraphics[scale=0.17]{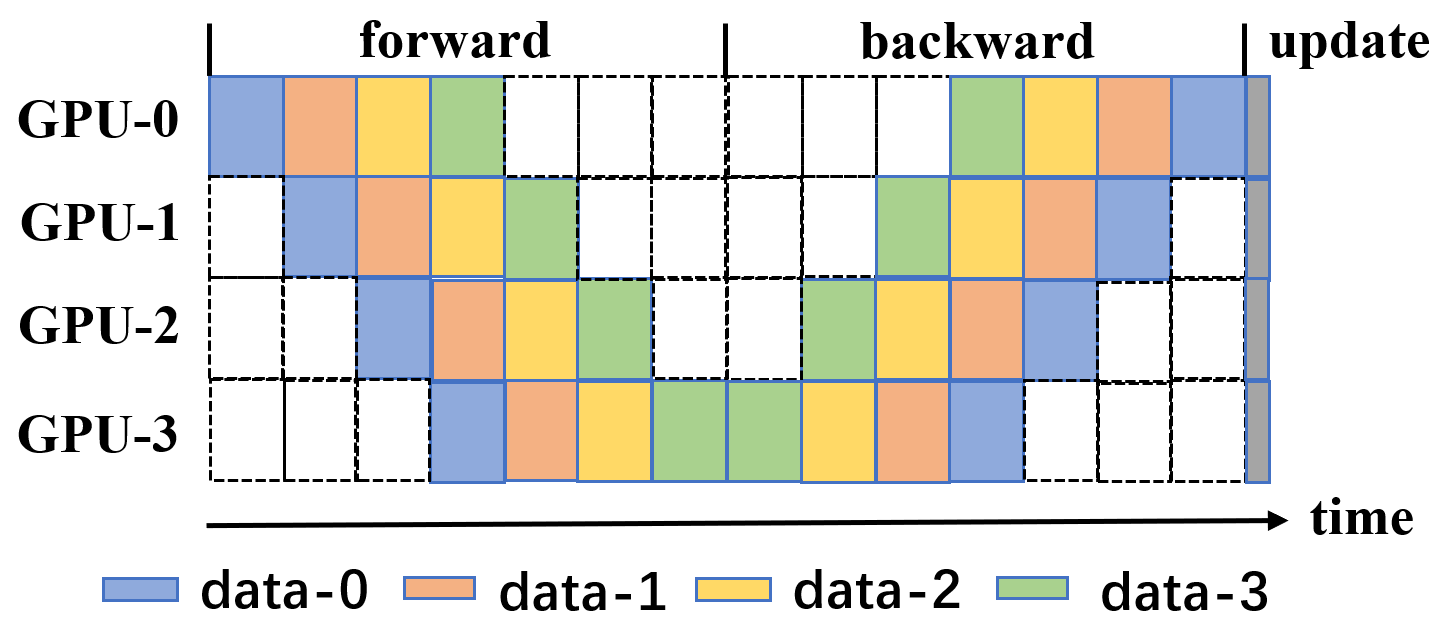}
\label{fig:pipeline}
\end{minipage}
}
\caption{Model-parallel and pipeline model-parallel methods.}    
\label{fig:parallel}    
\end{figure*}

\section{Pretraining Quantized LLMs Using Low-rank Structure}
\label{sec:pre-training}
\subsection{Pretraining Methods with Low-rank Structure}
\label{sec:pre-training_method}



Pretraining an LLM has high costs and long training time. We replace the linear layers in the LLM with low-rank structure to reduce the number of parameters. For example, the $\bm{W}^{K} \in \mathbb{R}^{4096 \times 4096}$ in Llama-7B can be replaced with two sub-linear layers, which have weight matrices of $\bm{A}^{K} \in \mathbb{R}^{4096 \times r}$ and $\bm{B}^{K} \in \mathbb{R}^{r \times 4096}$. The parameters of this layer are reduced by $\frac{4096}{2r}$. Using a low-rank structure, we do the pretraining of LLMs as follows.

1) We replace a linear layer of the transformer with two sub-linear layers, which use a hidden layer with a small width, as shown in Fig. \ref{fig:lowrank_training}. In particular, a weight matrix $\bm{W} \in \mathbb{R}^{n \times n}$ in Table \ref{table:parameters} is replaced with two matrices, $\bm{A}  \in \mathbb{R}^{n \times r}$ and $\bm{B}  \in \mathbb{R}^{r \times n}$. We refer to this as Method 1. 

2) We decompose the pretrained weight matrix $\bm{W} \in \mathbb{R}^{n \times n}$ in Table \ref{table:parameters} into low-rank matrices $\bm{A}  \in \mathbb{R}^{n \times r}$ and $\bm{B}  \in \mathbb{R}^{r \times n}$, as shown in Fig. \ref{fig:lowrank_training2}. Then, we use $\bm{A}$ and $\bm{B}$ to initialize a low-rank LLM. We refer to this as Method 2. 

3) Adding two narrow linear layers to the parallel path of the pretrained linear layer, we obtain the new linear layers as follows:
\begin{equation}
    \bm{y} = \alpha\bm{Wx}+(1-\alpha)\bm{BAx},~0 \leq \alpha \leq 1.
\end{equation}
The pretrained linear $\bm{W} \in \mathbb{R}^{n \times n}$ are frozen. At the beginning of pretraining, $\alpha$ is set close to $1$. Then, $\alpha$ is decreased during the pretraining, until it is $0$. The pretrained LLM has only the low-rank matrices $\bm{A}  \in \mathbb{R}^{n \times r}$ and $\bm{B}  \in \mathbb{R}^{r \times n}$.

\subsection{Optimization for the Pretraining Stage}
\label{section:train_from_scratch}

We use low-rank LLMs in the pretraining stage. There are some challenges, as mentioned in Section  \ref{sec:chaInTrain}, to be addressed.

\begin{table}
  \caption{GPU memory consumption of intermediate variables. }
  \centering
  \begin{tabular}{ccc}
\toprule
\textbf{variables} & \textbf{\#Parameters} & \textbf{Size (GB)}\\
\midrule
$\bm{x}^e$ & $16.8$ M & $\ll 0.1$ \\
$\Tilde{\bm{x}}^e$ & $536,9$ M& $1.1$\\
$\bm{k}$ & $536,9$  M& $1.1$  \\
$\bm{q}$ & $536,9$  M& $1.1$  \\
$\bm{v}$ & $536,9$  M& $1.1$ \\
$\bm{qk}^T$ & $17.2$ B& $34.4$  \\
$\bm{s}$ & $17.2$ B& $34.4$  \\
$\bm{x}^{o}$ & $536,9$  M& $1.1$ \\
$\Tilde{\bm{x}}^o$ & $536,9$  M& $1.1$\\
$\bm{x}^{u}$ & $1,4$ B& $2.8$\\
$\bm{x}^{g}$ & $1.4$ B& $2.8$\\
$\bm{x}^{d}$ & $536,9$ M& $\ll 0.1$\\
\bottomrule
\end{tabular}
\label{table:memofinter}
\end{table}

\textbf{Recomputing technique to avoid storing intermediate variables}: 
Intermediate variables are generated in the forward pass, used in backward pass, and then released. Table \ref{table:memofinter} shows the GPU memory consumption of intermediate variables during the pretraining of Llama2-7B. The dimension, number of heads, batch size, and sequence length are $4096$, $32$, $1$, and $4096$, respectively.

Compared $\bm{x}^e$ with $\bm{x}^d$, other intermediate variables have more GPU memory consumption. There are two reasons: 1) other variables are in the decoder layer, which has $32$ layers. 2) Some variables, $\bm{qk}^T$, $\bm{s}$, have the same size as $\bm{k}, \bm{q}, \bm{v}$ for each head. There are $h = 32$ heads for each decoder layer.

By avoiding storing intermediate variables and recomputing them when needed, the GPU memory consumption can be significantly reduced, at the cost of more computations. The process of this method is as follows:
\begin{enumerate}
    \item In forward pass, only the input, as $\bm{x}^{e}$, is stored.
    \item In backward pass, the GPU recomputes other variables using $\bm{x}^{e}$ for a decoder layer.
    \item Calculating the gradients of parameters in this decoder layer, the variables are deleted after obtaining gradients.
    \item Repeat steps 1) - 2) until all decoder layers calculate the gradients.
    \item Obtaining all gradients, and the parameters are updated.
\end{enumerate}
In this method, the peak memory consumption is the size of intermediate variables in one decoder layer. The GPU memory consumption is reduced from $81$ GB to $2.5$ GB, as in Table\ref{table:memofinter}, at a cost of computational workload for an extra forward pass. We know that the computational workload of forward pass is half of backward pass. Therefore, the GPU memory consumption of intermediate variables can be reduced by $97\%$ with a $33\%$ increase in computation.

As shown in Table \ref{table:memofinter}, these intermediate variables have different sizes. By selectively choosing which intermediate variables to recompute, there is a good balance between memory consumption and computation. For example, $\bm{qk}^T$ and $\bm{s}$ are much bigger than other variables. By only recomputing $\bm{qk}^T$ and $\bm{s}$, we can reduce the GPU memory consumption from $81$ GB to $12.2$ GB, at the cost of about $6\%$ increases in computation.

\textbf{Pipeline model-parallel method}. For pretraining an LLM on $N$ GPUs, the model-parallel method loads the parameters on different GPUs, as shown in Fig. \ref{fig:modelparallel}. The computing is performed sequentially on these GPUs, and each GPU has a utilization of $\frac{1}{N}$. In the pipeline model-parallel method, the input data is split into $M \geq 1$ mini-batches. As shown in Fig. \ref{fig:pipeline}, we input these mini-batches into GPU-0 to GPU-3 sequentially and perform the backward pass in a reverse order. Then, the parameters are updated. In this method, the GPU has a utilization of $\frac{M}{N+M-1}$, where $M \ge 1$.

\textbf{Weight decomposition in parallel}: Decomposing weight $\bm{W}$ at different linear layers is independent. For example, Llama2-7B has $32$ decoder layers, and each decoder layer has $7$ linear layers as shown in Fig. \ref{fig:transformer_architecture}. The decomposition stage has a parallelism of $224 = 7 \times 32$. In this method, we distribute the LLM decoder layers evenly across multiple GPUs. For example, the $\bm{W}^K$ in different decoder layers have the same size. In a single GPU, we batch the decomposition of these $\bm {W}^K$ in different layers. On different GPUs, we perform the decomposition of different weight matrices, such as $\bm {W}^Q$ and $\bm {W}^V$, in parallel.

\subsection{Optimization for the Inference Stage}
\label{section:Optimization_for_infer}

Utilizing the low-rank structure of the weights, the number of parameters of GPT2-1.5B can be reduced from $1.56$ B to $0.59$ B. This in turn can reduce the model size from $3.12$ GB to $1.18$ GB. By quantizing into $8$-bit or $4$-bit precision, the model will have a size of less than $0.59$ GB, and thus can be loaded in mobile phones. 

As described in Section \ref{sec:lowrank_structure}, the number of parameters of Llama2-7B can be reduced from $6.74$ B to $1.32$ B. For inputs and outputs that have $100$ tokens in the inference stage as in Section \ref{sec:challangeininfer}, our method can reduce the computational workload from $210$ to $40$ TFLOPS and thus reduce the response time from about $105$ to $20$ seconds.

\begin{figure*}[htbp]
\subfigure[full finetuning]{   
\begin{minipage}{9cm}
\includegraphics[scale=0.14]{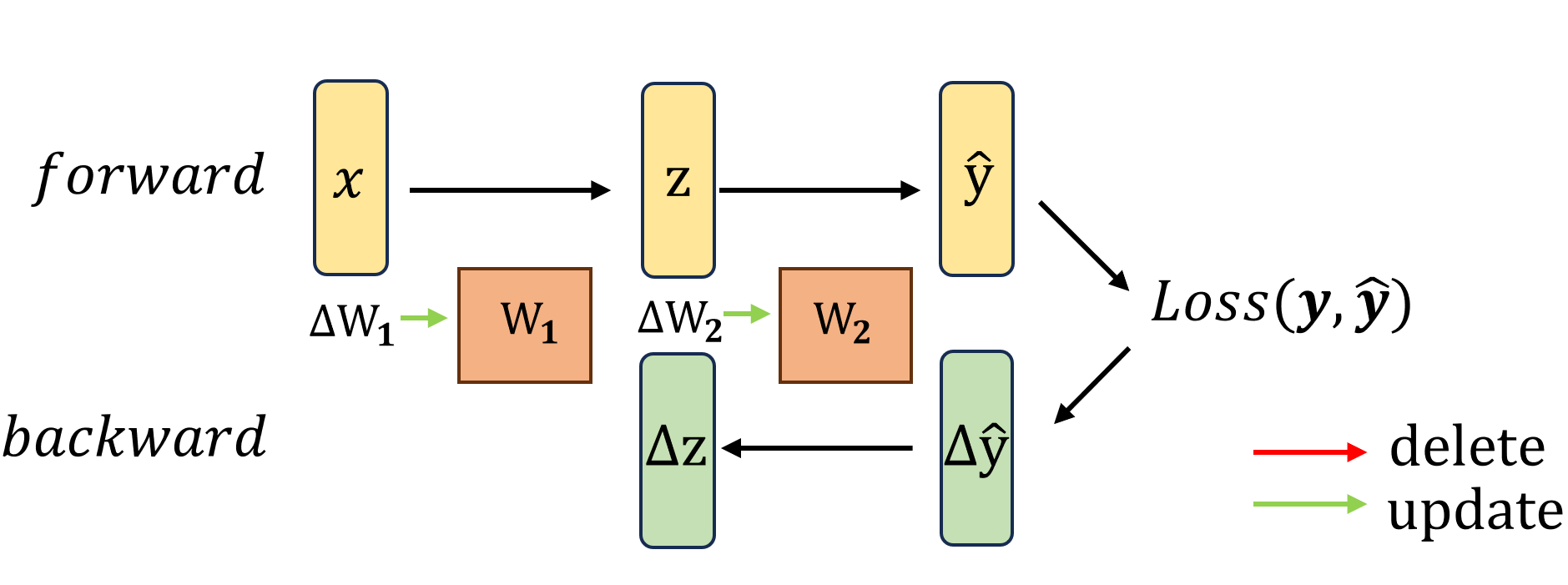}  
\label{fig:original_fb}
\end{minipage}
}
\subfigure[low-rank adaptation finetuning]{ 
\begin{minipage}{9cm}
\includegraphics[scale=0.15]{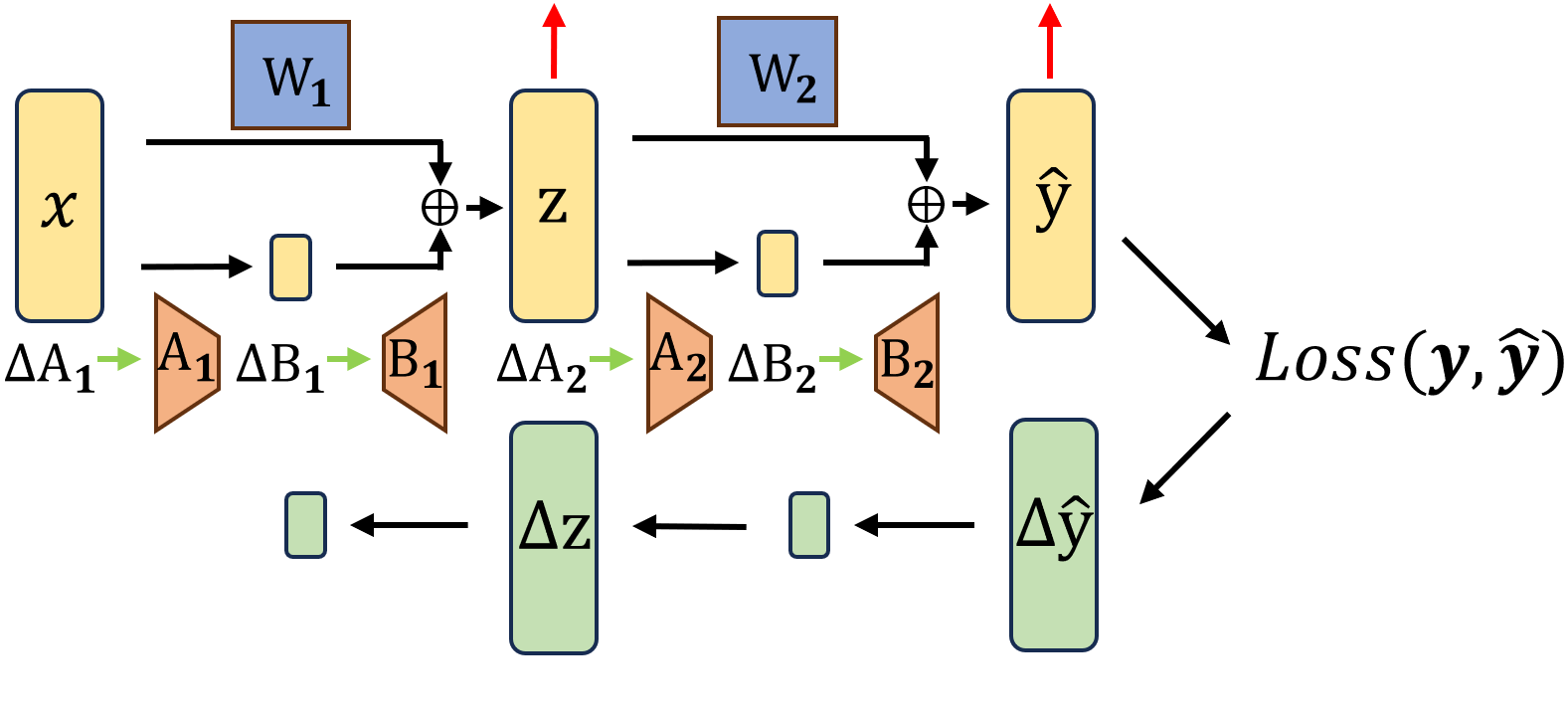}
\label{fig:lora_fb}
\end{minipage}
}
\caption{Finetuning flow of low-rank adaptation method.}    
\label{fig:flowoflowrank}    
\end{figure*}


\section{Finetuning Quantized LLMs Using Low-rank Structure}
\label{sec:fine-tuning}

\begin{figure}[b]
\centering
\includegraphics[width=9cm]{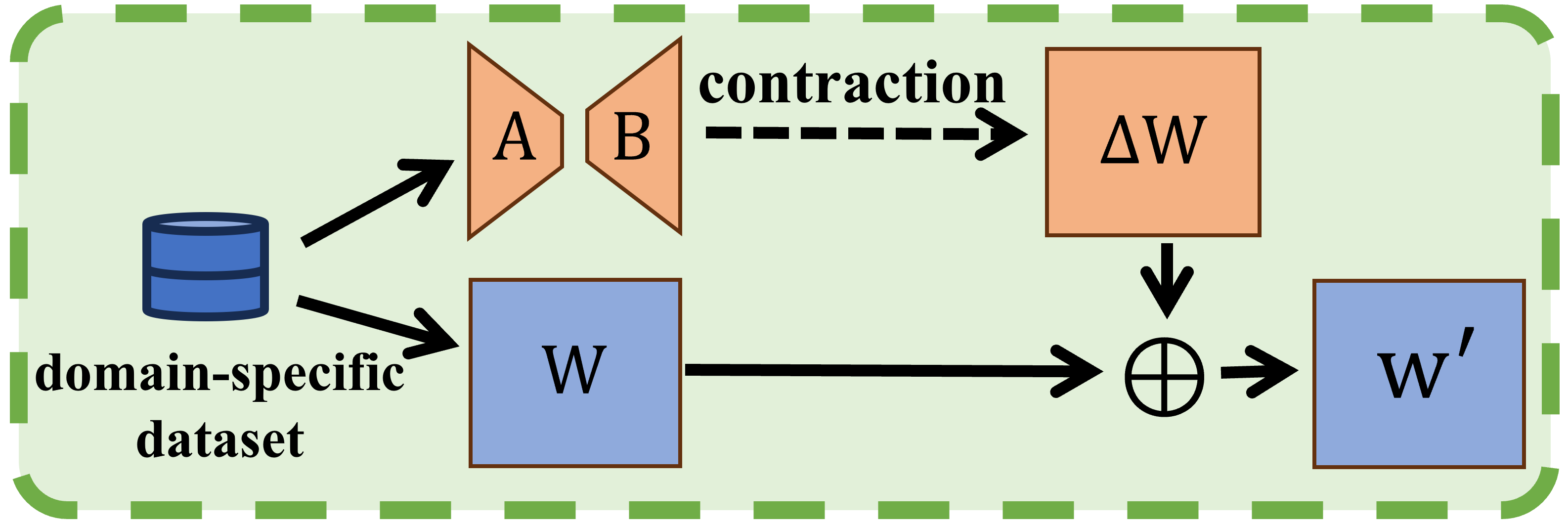}
\caption{The low-rank adaptation method for finetuning.}
\label{fig:low-rank adaptative finetuning}
\end{figure}

\subsection{Our Finetuning Method}
\label{sec:lora}
Many ongoing efforts are on finetuning pretrained models to obtain domain-specific LLMs. Compared with pre-training, fine-tuning of LLMs requires data of a much smaller scale. Low-rank adaptation (LoRA) method \cite{hulora} provides a low-cost finetuning solution for LLMs, as shown in Fig. \ref{fig:low-rank adaptative finetuning}. This method finetunes a pretrained LLM on a domain-specific dataset. The finetuned LLM gains knowledge of the specific domain.



 Given a pretrained LLM with weight $\bm{W}$, the LoRA method adapts it to  $\bm{W}^{'} = \bm{W} + \Delta \bm{W}$ by adding a low-rank $\Delta\bm{W} = \bm{B} \bm{A}$. The idea of the LoRA method \cite{hulora} is as follows:
\begin{equation}
\label{eq:new_linear_layer}
\bm{W}^{'}\bm{x} =(\bm{W}+\Delta\bm{W})\bm{x}= \bm{W}\bm{x}+\bm{B}\bm{A}\bm{x},
\end{equation}
where $\bm{W} \in \mathbb{R}^{n \times n}, \bm{A} \in \mathbb{R}^{r \times n}, \bm{B} \in \mathbb{R}^{n \times r}, r \ll n$. 


During the finetuning stage, the weight $\bm{W}$ is frozen, and the low-rank matrices $\bm{A}$ and $\bm{B}$ are trainable. In the inference stage, matrices product is performed with $\bm{A}$ and $\bm{B}$ and adding the result $\bm{\Delta W}$ to the pretrained weight $\bm{W}$, which is denoted as $\bm{W}'$.

\subsection{Optimization for the Finetuning Stage}




\begin{table}
\caption{The performance at different precisions.}
\Centering
\begin{tabular}{ccc}
\toprule
Precision & Performance (TFLOPS)\\
\midrule
FP32 & 19.5\\
FP16 & 312\\
int8 & 624\\
int4 & 1248\\
\bottomrule
\end{tabular}
\label{table:mm_time}
\end{table}

\textbf{Quantization}: The low-rank adaptation method significantly reduces the number of trainable parameters. However, the pretrained weights have a larger memory consumption. For example, the pretrained weights of Llama2-7B\cite{llama2} consume $14$ GB of GPU memory during finetuning. In our method, the pretrained weights are quantized to lower precision. Low-precision data has a smaller size and is better suited for computing using GPU tensor cores, as shown in Table \ref{table:mm_time}. The pretrained weights of LLMs are in matrix form. We quantize the weights on per-vector as follows:
\begin{itemize}
    \item The pretrained weight $\bm{W} \in \mathbb{R}^{n \times n}$ has $n$ vectors, calculating the difference between the maximum and minimum values to get the scaling factor of each vector.
    \item In each vector, there are $n$ elements multiplied with scaling factor. These multiplications are batched by mapping onto $n$ CUDA cores.
    \item Results are stored in low-precision formats in GPU memory, while the original weights are deleted.
\end{itemize}
In quantization, there is a trade-off between memory footprint and accuracy with \textsf{8-bit} and \textsf{8-bit} data formats. These two data formats can reduce the GPU memory footprint of the pretrained weights from $14$ GB to $8$ GB and $5$ GB, respectively.

\textbf{Intermediate variables:} As shown in Fig. \ref{fig:flowoflowrank}, the low-rank structure can reduce the trainable parameters from $\bm{W_1}$ to $\bm{A_1}$ and $\bm{B_1}$. However, the intermediate variables, such as $\bm{x}, \bm{z}$ and $\hat{\bm{y}}$, for low-rank finetuning have the same size as full finetuning. We recompute the intermediate variables as in Section \ref{section:train_from_scratch}.

\subsection{Optimization for the Inference Stage}

There is a growing interest in customized LLMs that are tailored to different scenarios or applications. In our method, we can view each customized model as a set of finetuned low-rank matrices $\bm{A}$ and $\bm{B}$, sharing the base pretrained model $\bm{W}$. The low-rank matrices have much fewer parameters, about $0.1\%$ of that of the base model. There are two benefits:
\begin{itemize}
    \item The storage can be significantly reduced for each customized LLM.
    \item By loading a base model with different finetuned low-rank matrices, the LLM can be used for different inference scenarios.
\end{itemize}

As described in Section \ref{sec:chaInInfer}, each generated token is added at the end of input sequence. Multiple tokens are calculated multiple times in inference stage. For example, the original input sequence has $100$ tokens and the LLM generates $10$ tokens. The original $100$ tokens are input into LLM $10$ times for inference. In our method, we store $\bm{K}, \bm{V} \in \mathbb{R}^{n \times l}$ of each layer, as shown in Fig. \ref{fig:transformer_architecture}. For each generated token, we input it into LLM to calculate $\hat{\bm{q}}, \hat{\bm{k}}, \hat{\bm{v}} \in \mathbb{R}^{n}$. Then concatenating $\hat{\bm{k}}, \hat{\bm{v}}$ with $\bm{K}, \bm{V}$, respectively. We cache the updated $\bm{K}, \bm{V} \in \mathbb{R}^{n \times (l+1)}$. Using this method, we can reduce the number of input tokens from $l$ to $1$, which can significantly reduce the computing workload in inference by $l \times$.

The cache of $\bm{K}, \bm{V}$ have a huge block of GPU memory, which varies with the length of input sequence. There is a memory fragmentation problem, leading to inefficient use of GPU memory. PagedAttention\cite{kwon2023efficient} provides a method that refers to the virtual memory technique, splitting the $\bm{K}, \bm{V}$ cache into multiple GPU blocks. There is a table mapping the logical cache and physical memory blocks.

\section{Performance Evaluation}
\label{section:performace}

\begin{figure*}[htbp]
\subfigure[Training loss for GPT2-127M]{   

\begin{minipage}{9cm}
\centering
\includegraphics[scale=0.21]{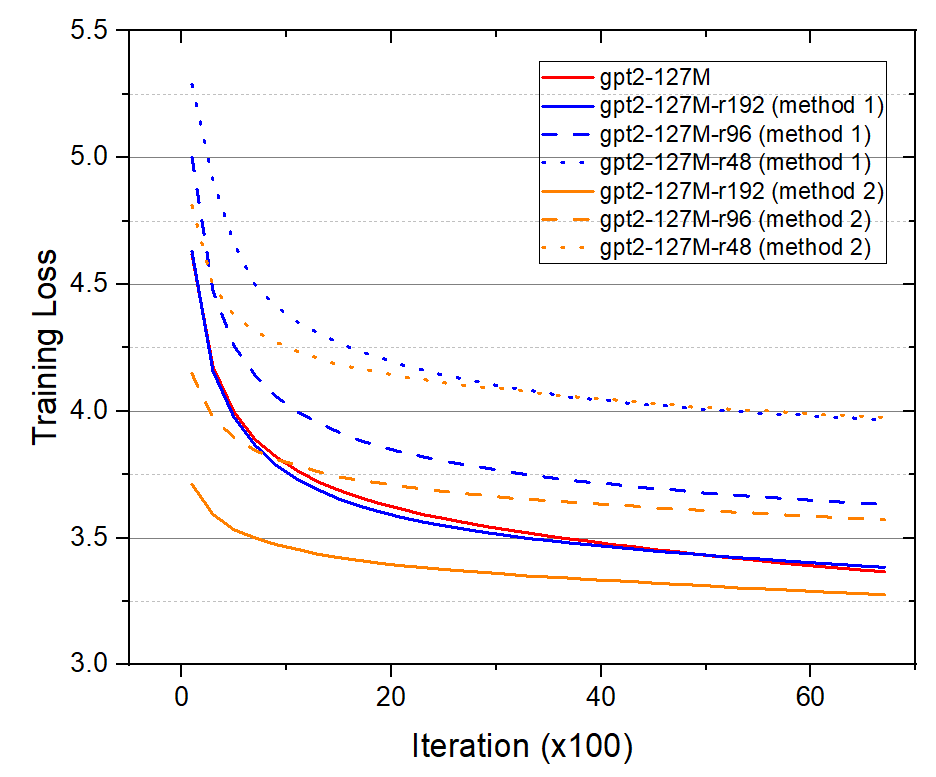}  
\label{fig:lossforGPT2}
\end{minipage}
}
\subfigure[Training loss for GPT2-1.5B]{ 
\begin{minipage}{9cm}
\centering
\includegraphics[scale=0.21]{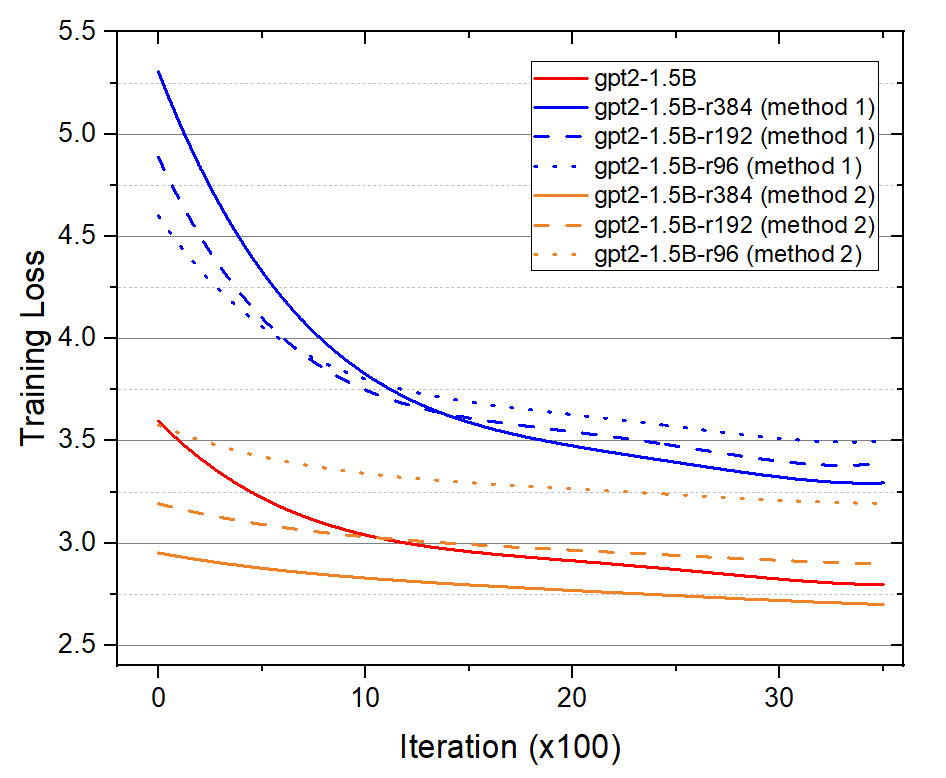}
\label{fig:lossforGPT2xl}
\end{minipage}
}
\caption{Training loss for pretraining GPT2-127M and GPT2-1.5B using method 1 and method 2.}    
\label{fig:pretrainingloss}    
\end{figure*}

\subsection{Experimental Settings}

\textbf{Server}: Our experiments were conducted on a DGX-2 server, equipped with two 64-core AMD EPYC 7742 CPUs, 8 NVIDIA A100 GPUs, and 2 TB of memory. The server was running Ubuntu 20.04 with CUDA 11.6, and we utilized PyTorch version 1.13 as our neural network framework.

\textbf{Datasets, Baselines, and Metrics}



\textbf{Training datasets.} We use the OpenWebText \cite{OpenWebText} dataset as our pretraining dataset. It is a clone of the GPT-2 WebText dataset and has about $11.2$ B tokens. We use the alpaca dataset \cite{alpaca} and FinGPT dataset \cite{2023finnlp} for our finetuning method. The alpaca dataset was constructed using the self-instruct method and the FinGPT dataset was collected from financial websites. It contains $52$K instruction-following examples. 

\textbf{Evaluation datasets.} We evaluate our methods on two types of tasks: general tasks and financial tasks. The general tasks used the following three datasets:
\begin{itemize}
    \item \textbf{BoolQ} \cite{clark2019boolq} is a question answering dataset. It has $16$K examples, and each example consists of a passage, question, and answer. The question is about the passage, and the answer is \textsf{yes} or \textsf{no}.
    \item \textbf{PIQA} \cite{Bisk2020piqa} is a dataset about physical commonsense reasoning. It has $21$K examples, and each example consists of a question and two possible solutions. The model to be evaluated needs to choose the most appropriate solution.
    \item \textbf{WinoGrande} \cite{ai2:winogrande} is a commonsense reasoning dataset. It has $44$K sentences, and each sentence has a blank to be filled. The model to be evaluated needs to choose the correct one from the given two options.
\end{itemize}
The financial tasks used the following three datasets:
\begin{itemize}
    \item \textbf{FPB} \cite{malo2014good} consists of $4.8$K sentences from financial news categorised by sentiment. Any news that may have a benefit or risk impact on investors is categorized as positive or negative. Other news that has no such effect is categorized as neutral.
    \item \textbf{FiQA SA} \cite{2018WWWFiQA} consists of $17$K sentences from microblog headlines and financial news. These sentences are classified according to sentiment as the FPB dataset.
    \item \textbf{TFNS} \cite{tfns} consists of $11.9$K sentences from annotated corpus of finance-related tweets. This dataset is used to classify finance-related tweets based on their sentiment.
    \item \textbf{NER} \cite{alvarado2015domain} consist of $1.4$K sentences from financial agreements. This dataset is used to recognize the named entity.
    
\end{itemize}

We use GPT2-127M\cite{GPT2}, GPT2-1.5B\cite{GPT2}, Llama2-7B \cite{llama2} and Llama2-13B \cite{llama2} as our base LLMs. 
\begin{itemize}
    \item GPT2-127M\cite{GPT2} has $127$M parameters. It contains $N = 12$ layers with $n = 786$ and $h = 12$. 
    \item GPT2-1.5B\cite{GPT2} has $1.5$Bparameters. It contains $N = 48$ layers with $n = 1600$ and $h = 25$. 
    \item Llama2-7B\cite{llama2} has $6.7$B parameters. It contains $N = 32$ layers with $n = 4096$ and $h = 32$.  
    \item Llama2-13B\cite{llama2} has $13$B parameters. It contains $N = 40$ layers with $n = 5120$ and $h = 40$.
\end{itemize}

We show the F1 weighted scores for these tasks.  For high-performance computing, we are interested in the following performance metrics:
\begin{itemize}
    \item \textbf{Trainable parameters}: We are interested in the number of trainable parameters. These parameters will vary during finetuning.

    \item \textbf{GPU memory footprint}: We accumulate the GPU memory usage of all GPUs used for finetuning.

    \item \textbf{Training Time}: Time defined as (the number of GPUs used) $\times$ (the process running time).    

\end{itemize}

\subsection{Results for Pretraining LLMs Using Low-rank Structures}





\begin{table*}
\caption{Result for pretraining using method 1 in Fig. \ref{fig:lowrankpretraining}.}
\Centering
\resizebox{\linewidth}{!}{
\begin{tabular}{c|ccc|ccc}
\toprule
Model \& Method & BoolQ & PIQA & WinoGrande & \#Params. (Model size)& Memory (GB) & Time (hours)\\
\midrule
GPT2-127M\cite{GPT2} & $\textbf{61.9}$ & $56.9$ &$48.9$ &$127.4$ M (127.4 MB)& $195.2$ & $138.7$\\
GPT2-127M-r192 & $ 48.1 $ & $ \textbf{57.7} $ & $ 50.7 $ & $ 77.5 $ M (77.5 MB)& $ 192.4 $ & $ 121.0 $\\
GPT2-127M-r96  & $ 38.0 $ & $ 56.7 $ & $ \textbf{51.3} $ & $ 58.5 $ M (58.5 MB)& $ 188.2 $ & $ 117.2 $\\
GPT2-127M-r48  & $ 51.8 $ & $ 55.8 $ & $ 50.0$ & $ 49.0 $ M (49.0 MB) & $ 187.6 $ & $ 116.4 $\\
\midrule
GPT2-1.5B\cite{GPT2} & $ \textbf{59.9} $ & $ \textbf{63.6} $ & $ 50.4 $ & $ 1.56 $ B (1.56 GB) & $ 301.6 $ & $ 546.4 $ \\
GPT2-1.5B-r384 & $ 58.5 $ & $ 58.7 $ & $ \textbf{50.6} $ & $ 0.59 $ B (0.59 GB)& $ 238.0 $ & $ 416.3 $ \\
GPT2-1.5B-r192 & $ 43.1 $ & $ 57.4 $ & $ 48.9 $ & $ 0.34 $ B (0.34 GB)& $ 235.3 $ & $ 395.6 $ \\
GPT2-1.5B-r96 & $ 43.7 $ & $ 57.2 $ & $ 49.5 $ & $ 0.21 $ B (0.21 GB)& $ 264.0 $ & $ 399.6 $ \\
\bottomrule
\end{tabular}
}
\label{table:result_initializing}
\end{table*}

\begin{table*}
\caption{Results for pretraining using method 2 in Fig. \ref{fig:lowrankpretraining}.}
\Centering
\resizebox{\linewidth}{!}{
\begin{tabular}{c|ccc|ccc}
\toprule
Model \& Method & BoolQ & PIQA & WinoGrande & \#Params. (Model size) & Memory (GB) & Time (hours)\\
\midrule
GPT2-127M\cite{GPT2} & $\textbf{61.9}$ & $56.9$ &$48.9$ &$127.4$ M (127.4 MB)& $195.2$ & $138.7$\\
GPT2-127M-r192 & $ 39.2 $ & $ \textbf{58.9} $ & $ 50.2 $ & $ 77.5 $ M (77.5 MB)& $ 172.4 $ & $ 120.1 $\\
GPT2-127M-r96  & $ 47.8 $ & $ 55.7 $ & $ \textbf{50.8} $ & $ 58.5 $ M (58.5 MB)& $ 196.1 $ & $ 116.9 $\\
GPT2-127M-r48  & $ 38.2 $ & $ 55.1 $ & $ 52.0 $ & $ 49.0 $ M (49.0 MB)& $ 155.4 $ & $ 115.9 $\\
\midrule
GPT2-1.5B\cite{GPT2} & $ 59.9 $ & $ 63.6 $ & $ 50.4 $ & $ 1.56 $ B (1.56 GB) & $ 301.6 $ & $ 546.4 $ \\
GPT2-1.5B-r384 & $ \textbf{60.7} $ & $ \textbf{65.9} $ & $ \textbf{55.4} $ & $ 0.59 $ B (0.59 GB) & $ 274.1 $ & $ 414.2 $ \\
GPT2-1.5B-r192 & $ 46.7 $ & $ 62.4 $ & $ 51.5 $ & $ 0.34 $ B (0.34 GB) & $ 275.8 $ & $ 401.9 $ \\
GPT2-1.5B-r96 & $ 40.3 $ & $ 60.0 $ & $ 51.9 $ & $ 0.21 $ B (0.21 GB) & $ 255.9 $ & $ 398.4 $ \\
\bottomrule
\end{tabular}
}
\label{table:result_decomposing}
\end{table*}

\begin{figure*}[htbp]
\subfigure[Training loss for Llama2-7B]{   
\begin{minipage}{9cm}
\centering
\includegraphics[scale=0.1]{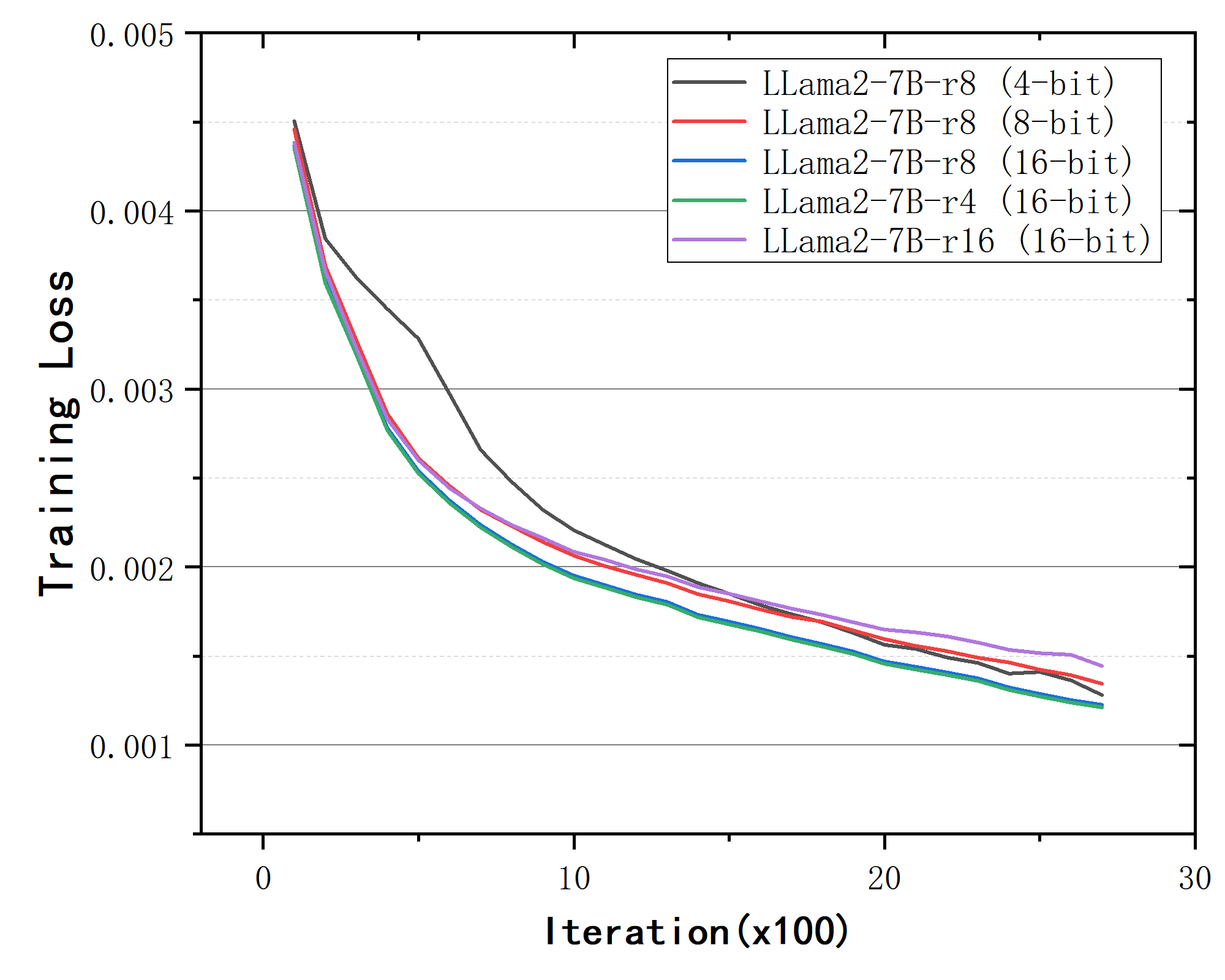}  
\label{fig:lossforLlama27b}
\end{minipage}
}
\subfigure[Training loss for Llama2-13B]{ 
\begin{minipage}{9cm}
\centering
\includegraphics[scale=0.1]{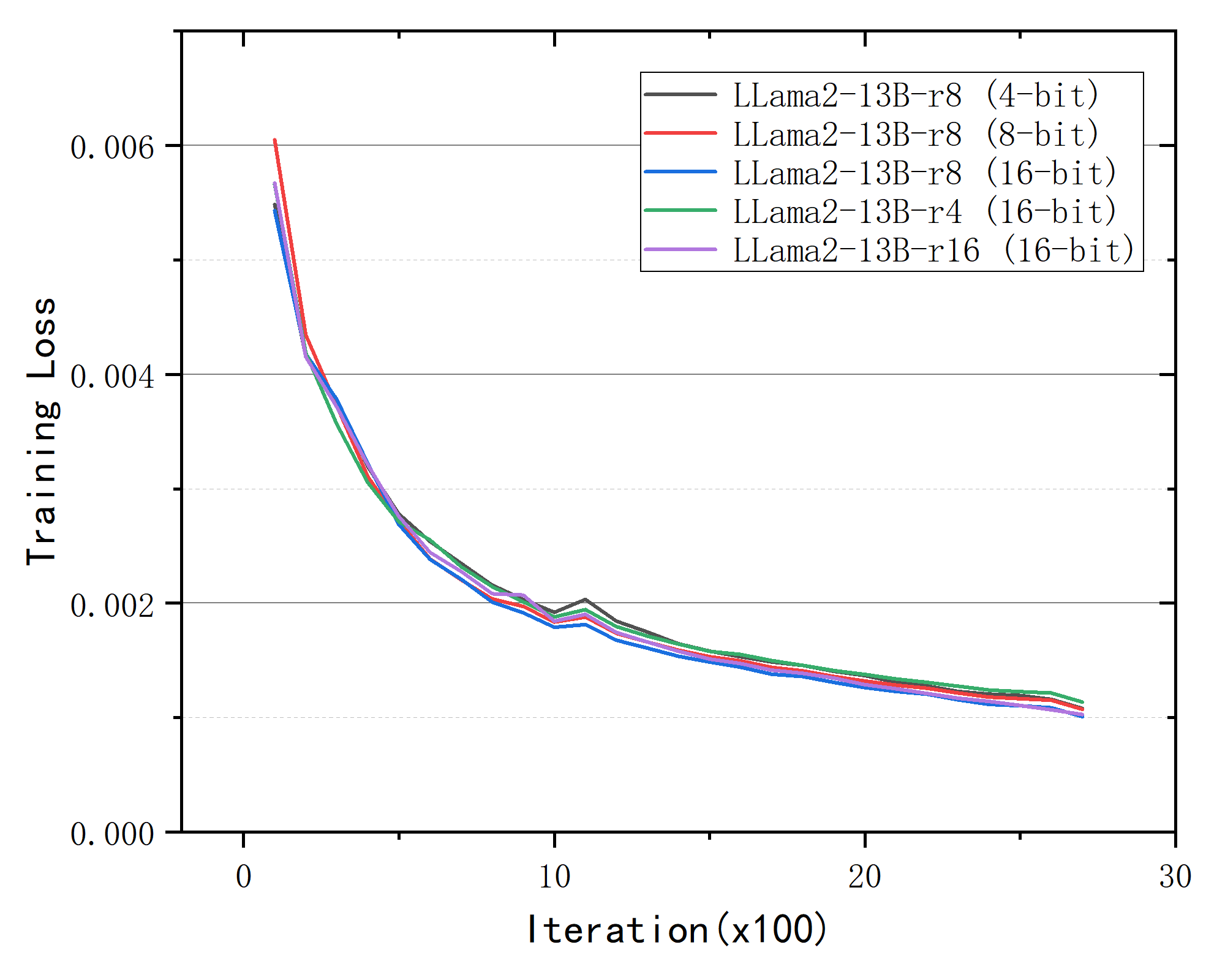}
\label{fig:lossforLlama213b}
\end{minipage}
}


\caption{Training loss for finetuning Llama2-7B, Llama2-13B with low-rank structure.}    
\label{fig:finetuningloss}  
\end{figure*}

\begin{table*}
\caption{Results for finetuning in Fig. \ref{fig:low-rank adaptative finetuning}.}
\Centering
\resizebox{\linewidth}{!}{
\begin{tabular}{c|ccc|cccc}
\toprule
Model \& Method & BoolQ & PIQA & WinoGrande & FPB & FiQA SA & TFNS\\
\midrule
Llama2-7B \cite{llama2} &$76.5$ & $79.8$ & $\textbf{70.1}$ & $51.8$ & $68.6$ & $80.9$ \\
Llama2-7B-r8 (4-bit) & $76.8$ & $77.0$ & $40.2$ & $73.8$ & $68.9$ & $70.2$\\
Llama2-7B-r8 (8-bit) & $78.9$ & $\textbf{81.8}$ & $59.1$ & $85.0$ & $86.0$ & $\textbf{89.4}$\\
Llama2-7B-r8 (16-bit)& $\textbf{82.8}$ & $79.1$ & $59.0$ & $85.6$ & $84.5$ & $89.4$\\
Llama2-7B-r4 (16-bit) & $81.4$ & $79.8$ & $56.9$ & $51.8$ & $68.6$ & $53.6$\\
Llama2-7B-r16 (16-bit) & $82.5$ & $80.1$ & $58.4$ & $\textbf{86.4}$ & $\textbf{86.8}$ & $89.1$\\

\midrule
Llama2-13B \cite{llama2}& $74.2$ & $72.3$ & $\textbf{67.7}$ & $54.2$ & $73.0$ & $63.1$\\
Llama2-13B-r8 (4-bit)& $84.6$ & $83.0$ & $63.0$ & $70.1$ & $69.2$ & $73.0$\\
Llama2-13B-r8 (8-bit)&$84.7$ & $83.2$ & $62.8$ & $\textbf{87.7}$ & $\textbf{88.6}$ & $90.2$\\
Llama2-13B-r8 (16-bit) &$85.0$ & $83.1$ & $62.8$ & $40.0$ & $56.1$ & $58.9$\\
Llama2-13B-r4 (16-bit) &$84.2$ & $82.9$ & $62.7$ & $26.4$ & $54.2$ & $39.9$\\
Llama2-13B-r16 (16-bit)& $\textbf{86.1}$ & $\textbf{83.9}$ & $63.0$ & $87.6$ & $84.2$ & $\textbf{90.4}$\\


\bottomrule
\end{tabular}
}

\label{table:llm_re3}
\end{table*}

We use the GPT2-127M\cite{GPT2} and GPT2-1.5B\cite{GPT2} as the base LLM. We replace all weight matrices in Table \ref{table:parameters} with low-rank structure, except for $\bm{W}^{E}$. For GPT2-127M, we set rank $r=192, 96, 48$, respectively; For GPT2-1.5B, we set rank $r=384, 192, 96$, respectively. We pretrain our LLMs on the OpenWebText dataset using method 1 and method 2 in Section \ref{sec:pre-training_method}. We evaluate them on general tasks. 

For method 1 and method 2 as in Section \ref{sec:pre-training_method}, the training losses over training iterations are shown in Fig. \ref{fig:pretrainingloss}. For GPT2-127M and GPT2-1.5B, our GPT2-127M-r192 (method 2) and GPT2-1.5B-r384 (method 2) have smaller loss values. For each model, the training loss of method 2 converges faster than method 1.

Table \ref{table:result_initializing} reports the testing accuracy, number of parameters, model size (8-bit), GPU memory footprint and training times for pretraining using method 1. For GPT2-127M, our method GPT2-127M-r192 achieves an accuracy increase of $0.8\%$ and $1.8\%$ on PIQA and WinoGrande, $1.64 \times$ compression ratio, and speedup of $1.14\times$ in training time; 
For GPT2-1.5B, our method GPT2-1.5B-r384 achieves an accuracy increase of $0.2\%$ on WinoGrande, $2.64 \times$ compression ratio, and speedup of $1.31\times$ in training time.

Table \ref{table:result_decomposing} reports the testing accuracy, number of parameters, model size (8-bit), GPU memory footprint and training times for pretraining using method 2. For GPT2-127M, our method GPT2-127M-r192 achieves an accuracy increase of $2.0\%$ and $1.3\%$ on PIQA and WinoGrande, $1.64 \times$ compression ratio, and speedup of $1.15\times$ in training time; 
For GPT2-1.5B, our method GPT2-1.5B-r384 achieves an accuracy increase of $0.8\%$, $1.5\%$, $5.0\%$ on these three datasets, $2.64 \times$ compression ratio, and speedup of $1.31\times$ in training time; Our method GPT2-1.5B-r96 achieves an accuracy increase of $1.5\%$ on WinoGrande, $7.43 \times$ compression ratio, and speedup of $1.37\times$ in training time.

\textbf{Inference performance}.
During the inference, our two methods have the same performance. 

For GPT2-127M\cite{GPT2}, our methods, GPT2-127M-r192, GPT2-127M-r96 and GPT2-127M-r48, achieve speedup of $1.2\times$ in inference time, from $0.06$ seconds to $0.05$ seconds. For GPT-1.5B, our methods, GPT2-1.5B-r384, GPT2-1.5B-r192, GPT2-1.5B-r96, achieve speedup of $1.9\times$ in inference time, from $0.43$ seconds to $0.23$.

\subsection{Results for Finetuning LLM Using Low-rank Structure}

\begin{table*}
\caption{Results for finetuning in Fig. \ref{fig:low-rank adaptative finetuning}.}
\Centering
\resizebox{\linewidth}{!}{
\begin{tabular}{c|ccc|cc}
\toprule
 \multirow{2}{*}{Model \& Method } & \multicolumn{3}{c|}{Finetuning} & \multicolumn{2}{c}{Inference} \\ \cline{2-6} & \#Params & Memory (GB) & Time (hours) & Model Size (GB)& Time (seconds)\\
\midrule
Llama2-7B \cite{llama2} & $6.7$ B & $112$ & $184,320$ & $13.4$ & $0.21$\\
Llama2-7B-r8 (4-bit) & $4.2 $ M  & $11.8$ & $7.8$ & $5.1$ & $0.23$ \\
Llama2-7B-r8 (8-bit) & $4.2 $ M  & $14.2$ & $4.7$ & $7.9$ & $0.13$\\
Llama2-7B-r8 (16-bit) & $ 4.2 $ M  & $20.3$ & $3.3$ & $13.4$ & $0.21$\\
Llama2-7B-r4 (16-bit) & $ 2.1 $ M  & $20.3$ & $4.0$  & $13.4$ & $0.21$\\
Llama2-7B-r16 (16-bit) &  $ 8.4 $ M  & $24.3$ & $2.5$ &  $13.4$ & $0.21$\\

\midrule
Llama2-13B \cite{llama2} &  $6.7$ B & $208$ & $368,640$ & $26.2$ & $0.39$ \\
Llama2-13B-r8 (4-bit) &  $9.8 $ M  & $17.7$ & $8.5$ & $9.8$ & $0.78$\\
Llama2-13B-r8 (8-bit) &  $9.8 $ M  & $30.4$ & $6.8$ & $15.5$ & $0.23$\\
Llama2-13B-r8 (16-bit) &$9.8 $ M &  $32.7$ & $5.4$ & $26.2$ & $0.39$ \\
Llama2-13B-r4 (16-bit) &  $ 4.9 $ M  & $32.7$ & $5.4$ & $26.2$ & $0.39$\\
Llama2-13B-r16 (16-bit) &  $ 19.6 $ M  & $33.0$ & $6.1$ & $26.2$ & $0.39$\\


\bottomrule
\end{tabular}
}
\label{table:llm_re4}
\end{table*}

We finetune Llama2-7B\cite{llama2} and Llama2-13B\cite{llama2}. The pretrained weights have a data precision of $16$-bit. We finetune these LLMs with low-rank structures with rank$=4, 8, 16$. For rank = $8$, we quantize these pretrained weights in $4$-bit and $8$-bit data precision, respectively.

Fig. \ref{fig:finetuningloss} shows the training losses over iterations for our finetuning methods. For both Llama2-7B and Llama2-13B, these methods have similar loss curves. For the Llama2-7B, the Llama2-7B-r8 (16-bit) method has the smallest loss. For the Llama2-13B, the Llama2-7B-r8 (16-bit) method has the smallest loss. 

Table \ref{table:llm_re3} reports the testing results on general tasks and financial tasks. For Llama2-7B, our methods Llama2-7B-r8 (8-bit) and Llama2-7B-r16 (16-bit) achieve average accuracy increase of $8.7\%$ and $11.8\%$, respectively; For Llama2-13B, our methods Llama2-13B-r8 (8-bit) and Llama2-13B-r16 (16-bit) achieve average accuracy increase of $15.5\%$ and $15.1\%$, respectively.

Table \ref{table:llm_re4} reports the number of trainable parameters, GPU memory footprint and finetuning time. For Llama2-7B, our methods Llama2-7B-r8 (4-bit) and Llama2-7B-r8 (8-bit) achieve a memory compression ratio of $9.5\times$ and $7.8\times$, reduce the model size from $13.4$ GB to $5.1$ GB and $7.9$ GB, respectively; For Llama2-13B, our method Llama2-13B-r8 (4-bit) and Llama2-7B-r8 (8-bit) achieve a memory compression ratio of $11.8\times$ and $6.8\times$, reduce the model size from $26.2$ GB to $9.8$ GB and $15.5$ GB, respectively.

\section{Conclusion and Future Work}
\label{section:ConclusionandFutureWork}

In this paper, we proposed an efficient pretraining and finetuning method for financial LLMs with low-rank structure. By taking advantage of the intrinsic low-rank dimension of LLMs, we replaced the linear layer in the transformer structure with low-rank matrices. For pretraining, we presented two methods: directly pretraining a low-rank LLM and decomposing the pretrained weights with low-rank structure to initialize a low-rank LLM. For finetuning, we added two low-rank matrices to the parallel path of the pretrained linear layer. The pretrained weights are frozen and only the low-rank weights are trainable. To further reduce the GPU memory consumption, we quantized the pretraining weights during the finetuning. We tested the proposed low-rank structure for pretraining GPT-127M and GPT-1.5B, and for finetuning Llama2-7B and Llama2-13B. The results show that we can achieve a high model performance with improved memory reduction and training speedup. 


In the future, we will extend the low-rank structure to larger LLMs, e.g., Llama2-70B. We plan to implement a library that can build LLMs with low-rank structure.

\bibliographystyle{elsarticle-num}


\begin{thebibliography}{10}
\expandafter\ifx\csname url\endcsname\relax
  \def\url#1{\texttt{#1}}\fi
\expandafter\ifx\csname urlprefix\endcsname\relax\def\urlprefix{URL }\fi
\expandafter\ifx\csname href\endcsname\relax
  \def\href#1#2{#2} \def\path#1{#1}\fi

\bibitem{vaswani2017attention}
A.~Vaswani, N.~Shazeer, N.~Parmar, J.~Uszkoreit, L.~Jones, A.~N. Gomez,
  {\L}.~Kaiser, I.~Polosukhin, Attention is all you need, Advances in neural
  information processing systems (2017) 30 (2017).

\bibitem{llama2}
H.~Touvron, L.~Martin, K.~Stone, P.~Albert, A.~Almahairi, Y.~Babaei,
  N.~Bashlykov, S.~Batra, P.~Bhargava, S.~Bhosale, et~al., Llama 2: Open
  foundation and fine-tuned chat models, arXiv preprint arXiv:2307.09288
  (2023).

\bibitem{krizhevsky2014one}
A.~Krizhevsky, One weird trick for parallelizing convolutional neural networks,
  arXiv preprint arXiv:1404.5997 (2014).

\bibitem{hulora}
E.~J. Hu, P.~Wallis, Z.~Allen-Zhu, Y.~Li, S.~Wang, L.~Wang, W.~Chen, et~al.,
  {LoRA}: Low-rank adaptation of large language models, in: International
  Conference on Learning Representations, 2022.

\bibitem{dettmers2023qlora}
T.~Dettmers, A.~Pagnoni, A.~Holtzman, L.~Zettlemoyer, Qlora: Efficient
  finetuning of quantized llms, arXiv preprint arXiv:2305.14314 (2023).

\bibitem{xu2023tensorgpt}
M.~Xu, Y.~L. Xu, D.~P. Mandic, Tensor{GPT}: Efficient compression of the
  embedding layer in {LLM}s based on the tensor-train decomposition, arXiv
  preprint arXiv:2307.00526 (2023).

\bibitem{GPT3}
T.~Brown, B.~Mann, N.~Ryder, M.~Subbiah, J.~D. Kaplan, P.~Dhariwal,
  A.~Neelakantan, P.~Shyam, G.~Sastry, A.~Askell, et~al., Language models are
  few-shot learners, Advances in Neural Information Processing Systems 33
  (2020) 1877--1901.

\bibitem{falcon}
E.~Almazrouei, H.~Alobeidli, A.~Alshamsi, A.~Cappelli, R.~Cojocaru, M.~Debbah,
  E.~Goffinet, D.~Heslow, J.~Launay, Q.~Malartic, B.~Noune, B.~Pannier,
  G.~Penedo, {Falcon-40B}: an open large language model with state-of-the-art
  performance (2023).

\bibitem{du2022glm}
Z.~Du, Y.~Qian, X.~Liu, M.~Ding, J.~Qiu, Z.~Yang, J.~Tang, Glm: General
  language model pretraining with autoregressive blank infilling, in:
  Proceedings of the 60th Annual Meeting of the Association for Computational
  Linguistics (Volume 1: Long Papers), 2022, pp. 320--335.

\bibitem{jiang2023mistral}
A.~Q. Jiang, A.~Sablayrolles, A.~Mensch, C.~Bamford, D.~S. Chaplot, D.~d.~l.
  Casas, F.~Bressand, G.~Lengyel, G.~Lample, L.~Saulnier, et~al., Mistral 7{B},
  arXiv preprint arXiv:2310.06825 (2023).

\bibitem{GPT2}
A.~Radford, J.~Wu, R.~Child, D.~Luan, D.~Amodei, I.~Sutskever, et~al., Language
  models are unsupervised multitask learners, OpenAI blog 1~(8) (2019) 9.

\bibitem{GPT1}
A.~Radford, K.~Narasimhan, T.~Salimans, I.~Sutskever, et~al., Improving
  language understanding by generative pre-training (2018).

\bibitem{instructGPT}
L.~Ouyang, J.~Wu, X.~Jiang, D.~Almeida, C.~Wainwright, P.~Mishkin, C.~Zhang,
  S.~Agarwal, K.~Slama, A.~Ray, et~al., Training language models to follow
  instructions with human feedback, Advances in Neural Information Processing
  Systems 35 (2022) 27730--27744.

\bibitem{touvron2023llama}
H.~Touvron, T.~Lavril, G.~Izacard, X.~Martinet, M.-A. Lachaux, T.~Lacroix,
  B.~Rozi{\`e}re, N.~Goyal, E.~Hambro, F.~Azhar, et~al., {LLaMA}: Open and
  efficient foundation language models, arXiv preprint arXiv:2302.13971 (2023).

\bibitem{kaplan2020scaling}
J.~Kaplan, S.~McCandlish, T.~Henighan, T.~B. Brown, B.~Chess, R.~Child,
  S.~Gray, A.~Radford, J.~Wu, D.~Amodei, Scaling laws for neural language
  models, arXiv preprint arXiv:2001.08361 (2020).

\bibitem{cobbe2021training}
K.~Cobbe, V.~Kosaraju, M.~Bavarian, M.~Chen, H.~Jun, L.~Kaiser, M.~Plappert,
  J.~Tworek, J.~Hilton, R.~Nakano, et~al., Training verifiers to solve math
  word problems, arXiv preprint arXiv:2110.14168 (2021).

\bibitem{wei2022emergent}
J.~Wei, Y.~Tay, R.~Bommasani, C.~Raffel, B.~Zoph, S.~Borgeaud, D.~Yogatama,
  M.~Bosma, D.~Zhou, D.~Metzler, et~al., Emergent abilities of large language
  models, arXiv preprint arXiv:2206.07682 (2022).

\bibitem{novikov2015tensorizing}
A.~Novikov, D.~Podoprikhin, A.~Osokin, D.~P. Vetrov, Tensorizing neural
  networks, Advances in Neural Information Processing Systems 28 (2015).

\bibitem{hayashi2019exploring}
K.~Hayashi, T.~Yamaguchi, Y.~Sugawara, S.-i. Maeda, Exploring unexplored tensor
  network decompositions for convolutional neural networks, Advances in Neural
  Information Processing Systems 32 (2019).

\bibitem{zhang2019cutensor}
T.~Zhang, X.-Y. Liu, cutensor-tubal: Optimized gpu library for low-tubal-rank
  tensors, in: ICASSP 2019-2019 IEEE International Conference on Acoustics,
  Speech and Signal Processing (ICASSP), IEEE, 2019, pp. 8583--8587.

\bibitem{lu2019high}
H.~Lu, T.~Zhang, X.-Y. Liu, High-performance homomorphic matrix completion on
  gpus, in: 2019 IEEE 21st International Conference on High Performance
  Computing and Communications; IEEE 17th International Conference on Smart
  City; IEEE 5th International Conference on Data Science and Systems
  (HPCC/SmartCity/DSS), IEEE, 2019, pp. 1627--1634.

\bibitem{li2019high}
H.~Li, T.~Zhang, R.~Zhang, X.-Y. Liu, High-performance tensor decoder on gpus
  for wireless camera networks in iot, in: 2019 IEEE 21st International
  Conference on High Performance Computing and Communications; IEEE 17th
  International Conference on Smart City; IEEE 5th International Conference on
  Data Science and Systems (HPCC/SmartCity/DSS), IEEE, 2019, pp. 1619--1626.

\bibitem{zhang2021high}
T.~Zhang, W.~Kan, X.-Y. Liu, High performance gpu primitives for graph-tensor
  learning operations, Journal of Parallel and Distributed Computing 148 (2021)
  125--137.

\bibitem{zhang2019cutensor-tubal}
T.~Zhang, X.-Y. Liu, X.~Wang, A.~Walid, cutensor-tubal: Efficient primitives
  for tubal-rank tensor learning operations on gpus, IEEE Transactions on
  Parallel and Distributed Systems 31~(3) (2019) 595--610.

\bibitem{zhang2020high}
T.~Zhang, X.-Y. Liu, X.~Wang, High performance gpu tensor completion with
  tubal-sampling pattern, IEEE Transactions on Parallel and Distributed Systems
  31~(7) (2020) 1724--1739.

\bibitem{HT_tensor}
H.~Huang, X.-Y. Liu, W.~Tong, T.~Zhang, A.~Walid, X.~Wang, High performance
  hierarchical tucker tensor learning using gpu tensor cores, IEEE Transactions
  on Computers (2022).

\bibitem{CP_tensor}
X.-Y. Liu, Z.~Zhang, Z.~Wang, H.~Lu, X.~Wang, A.~Walid, High-performance tensor
  learning primitives using gpu tensor cores, IEEE Transactions on Computers
  (2022).

\bibitem{dettmers2022gpt3}
T.~Dettmers, M.~Lewis, Y.~Belkada, L.~Zettlemoyer, Gpt3. int8 (): 8-bit matrix
  multiplication for transformers at scale, Advances in Neural Information
  Processing Systems 35 (2022) 30318--30332.

\bibitem{llmint8}
T.~Dettmers, M.~Lewis, Y.~Belkada, L.~Zettlemoyer, Llm. int8 (): 8-bit matrix
  multiplication for transformers at scale, arXiv preprint arXiv:2208.07339
  (2022).

\bibitem{xiao2023smoothquant}
G.~Xiao, J.~Lin, M.~Seznec, H.~Wu, J.~Demouth, S.~Han, Smoothquant: Accurate
  and efficient post-training quantization for large language models, in:
  International Conference on Machine Learning, PMLR, 2023, pp. 38087--38099.

\bibitem{frantar2022gptq}
E.~Frantar, S.~Ashkboos, T.~Hoefler, D.~Alistarh, Gptq: Accurate post-training
  quantization for generative pre-trained transformers, arXiv preprint
  arXiv:2210.17323 (2022).

\bibitem{RoPE}
J.~Su, M.~Ahmed, Y.~Lu, S.~Pan, W.~Bo, Y.~Liu, Roformer: Enhanced transformer
  with rotary position embedding, Neurocomputing 568 (2024) 127063.

\bibitem{cheng2015exploration}
Y.~Cheng, F.~X. Yu, R.~S. Feris, S.~Kumar, A.~Choudhary, S.-F. Chang, An
  exploration of parameter redundancy in deep networks with circulant
  projections, in: Proceedings of the IEEE International Conference on Computer
  Vision, 2015, pp. 2857--2865.

\bibitem{kwon2023efficient}
W.~Kwon, Z.~Li, S.~Zhuang, Y.~Sheng, L.~Zheng, C.~H. Yu, J.~Gonzalez, H.~Zhang,
  I.~Stoica, Efficient memory management for large language model serving with
  pagedattention, in: Proceedings of the 29th Symposium on Operating Systems
  Principles, 2023, pp. 611--626.

\bibitem{OpenWebText}
A.~Gokaslan, V.~Cohen, Openwebtext,
  \url{https://github.com/Skylion007/openwebtext} (2019).

\bibitem{alpaca}
R.~Taori, I.~Gulrajani, T.~Zhang, Y.~Dubois, X.~Li, C.~Guestrin, P.~Liang,
  T.~B. Hashimoto, Stanford alpaca: An instruction-following llama model
  (2023).

\bibitem{2023finnlp}
X.-Y. Liu, G.~Wang, H.~Yang, D.~Zha, Data-centric {FinGPT}: Democratizing
  internet-scale data for financial large language models, NeurIPS Workshop on
  Instruction Tuning and Instruction Following (2023).

\bibitem{clark2019boolq}
C.~Clark, K.~Lee, M.-W. Chang, T.~Kwiatkowski, M.~Collins, K.~Toutanova, Boolq:
  Exploring the surprising difficulty of natural yes/no questions, in: NAACL,
  2019.

\bibitem{Bisk2020piqa}
Y.~Bisk, R.~Zellers, R.~L. Bras, J.~Gao, Y.~Choi, Piqa: Reasoning about
  physical commonsense in natural language, in: Thirty-Fourth AAAI Conference
  on Artificial Intelligence, 2020.

\bibitem{ai2:winogrande}
Winogrande: An adversarial winograd schema challenge at scale, 2019.

\bibitem{malo2014good}
P.~Malo, A.~Sinha, P.~Korhonen, J.~Wallenius, P.~Takala, Good debt or bad debt:
  Detecting semantic orientations in economic texts, Journal of the Association
  for Information Science and Technology 65~(4) (2014) 782--796.

\bibitem{2018WWWFiQA}
M.~Maia, S.~Handschuh, A.~Freitas, B.~Davis, A.~Balahur, Www'18 open challenge:
  Financial opinion mining and question answering, in: Companion of the The Web
  Conference 2018, 2018.

\bibitem{tfns}
N.~Magic, Twitter financial news senti-ment,
  \url{http://precog.iiitd.edu.in/people/anupama}, (2022).

\bibitem{alvarado2015domain}
J.~C.~S. Alvarado, K.~Verspoor, T.~Baldwin, Domain adaption of named entity
  recognition to support credit risk assessment, in: Proceedings of the
  Australasian Language Technology Association Workshop 2015, 2015, pp. 84--90.

\end{thebibliography}




\end{document}